\documentclass[11pt]{article}

\usepackage[preprint]{acl}
\usepackage{times}
\usepackage{latexsym}
\usepackage[T1]{fontenc}
\usepackage[utf8]{inputenc}
\usepackage{amssymb}
\usepackage{microtype}
\usepackage{inconsolata}

\usepackage{amsmath,amsfonts,bm}

\def\eqref#1{equation~\ref{#1}}

\def\1{\bm{1}}

\DeclareMathAlphabet{\mathsfit}{\encodingdefault}{\sfdefault}{m}{sl}
\SetMathAlphabet{\mathsfit}{bold}{\encodingdefault}{\sfdefault}{bx}{n}

\usepackage{url}
\usepackage{graphicx}
\usepackage{subcaption}
\usepackage{booktabs}
\usepackage{multirow}
\usepackage{tabularx}
\usepackage[table]{xcolor}
\usepackage{placeins}
\usepackage{fvextra}
\usepackage{float}
\usepackage{stfloats}
\definecolor{oursgray}{RGB}{242,245,247}
\definecolor{gaincolor}{RGB}{32,120,75}
\definecolor{losscolor}{RGB}{176,62,55}
\definecolor{samecolor}{RGB}{105,112,120}

\title{From Trajectories to Differentiated Skills: Cognitive Differentiation for Self-Improving Agents}

\author{%
  Yewei~Zhao\textsuperscript{1}\quad
  Derong~Xu\textsuperscript{1,3}\quad
  Wei~Chen\textsuperscript{1}\quad
  Yin~Xu\textsuperscript{1,*}\quad
  Peilun~Zhou\textsuperscript{2,*}\\
  \bfseries Xiaohan~Wang\textsuperscript{2}\quad
  Guojun~Yin\textsuperscript{2}\quad
  Wei~Lin\textsuperscript{2}\quad
  Mingjun~Xiao\textsuperscript{1,*}\\[0.5em]
  {\normalfont\normalsize\textsuperscript{1}University of Science and Technology of China}\\
  {\normalfont\normalsize\textsuperscript{2}Meituan\qquad
   \textsuperscript{3}City University of Hong Kong}\\[0.3em]
  {\normalfont\normalsize\textsuperscript{*}Corresponding authors.}%
}
\hypersetup{%
  pdftitle={From Trajectories to Differentiated Skills: Cognitive Differentiation for Self-Improving Agents},
  pdfauthor={Yewei Zhao, Derong Xu, Wei Chen, Yin Xu, Peilun Zhou, Xiaohan Wang, Guojun Yin, Wei Lin, Mingjun Xiao}%
}

\begin{document}

\maketitle

\begin{abstract}
Large language model (LLM) agents tackle complex tasks through reasoning and interaction with external environments. Agent skills provide persistent, transferable capabilities that guide task planning, inform contextual decisions, support tool use, and enable executable procedures. Yet existing skill-evolution methods largely organize knowledge around skill artifacts, overlooking the functional heterogeneity within individual skills and its implications for knowledge maintenance and use. Inspired by cognitive science, we propose Cognitive Differentiation in Skill Evolution (CDSE), which evolves a unified skill through four functionally distinct yet coordinated components: Meta, Experience, Tool, and Script. CDSE extracts evidence-grounded semantic gradients from successful and failed trajectories, consolidates them by functional role, and applies component-specific updates and validation with cross-component coordination and adaptive invocation. An analysis of 500 top-ranked skill packages shows that 96.8\% contain knowledge spanning at least two functional roles. Across six benchmarks, three agent models, and two execution harnesses, CDSE achieves the highest scores in all 36 settings, improving over the Base agent by 7.6 percentage points and the strongest competing method in each setting by 2.6 points on average. Ablation studies further support the importance of functional differentiation and adaptive invocation.
\end{abstract}

\section{Introduction}

\begin{figure}[!t]
  \centering
  \captionsetup[subfigure]{font=10pt,skip=3pt}
  \begin{subfigure}{\columnwidth}
    \centering
    \includegraphics[width=0.77\linewidth]{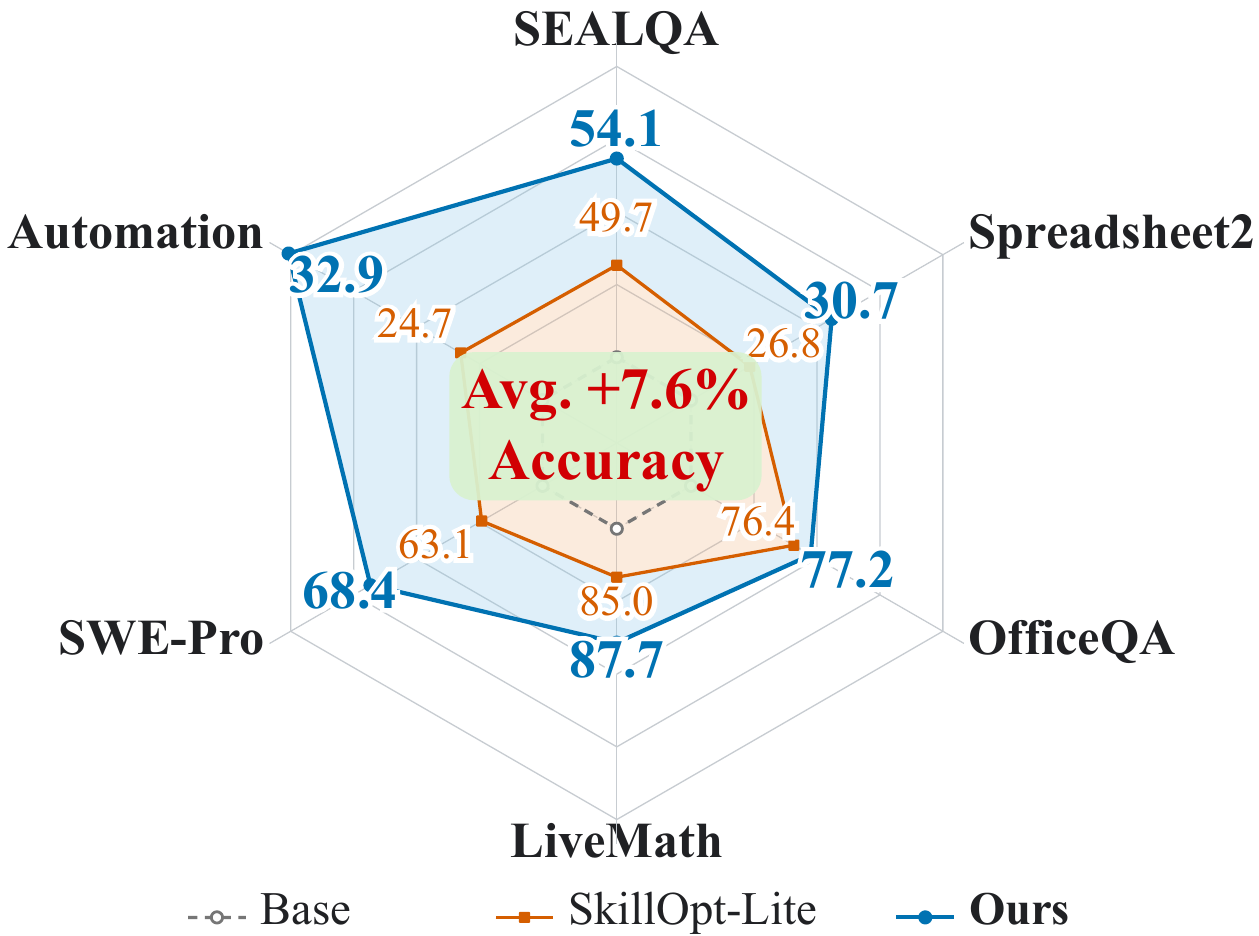}
    \caption{Task scores (\%) on six benchmarks for GLM-5.2 under Claude Code.}
    \label{fig:intro-results-comparison}
  \end{subfigure}
  \par\medskip
  \begin{subfigure}{\columnwidth}
    \centering
    \includegraphics[width=\linewidth]{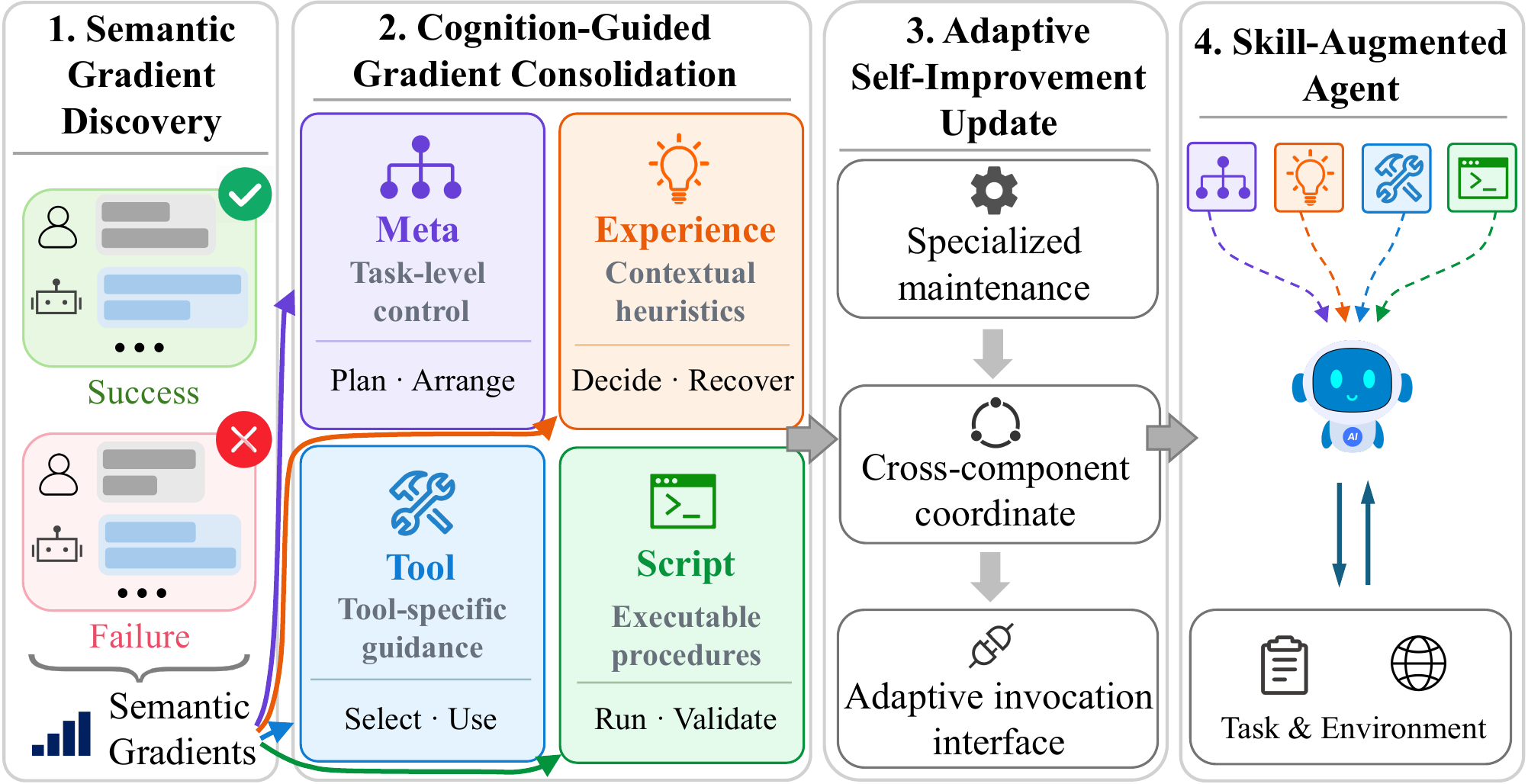}
    \caption{CDSE overview.}
    \label{fig:intro-cdse-overview}
  \end{subfigure}
  \caption{Performance and framework of CDSE.}
  \label{fig:intro-cdse}
\end{figure}

Large language model (LLM) agents tackle complex tasks through reasoning, planning, and interaction with external environments \citep{yao2023react,wang2024agentsurvey,yang2024sweagent}. External skills extend these capabilities by providing a persistent, reusable knowledge state without modifying model parameters \citep{wang2023voyager,wang2025awm,zheng2025skillweaver}. By leveraging feedback from evaluated execution trajectories, agents can continuously refine their skills to preserve effective behaviors, address recurring failures, and improve subsequent task performance. This has motivated growing research on trajectory-driven skill evolution as a means of continual agent self-improvement.

Recent studies have explored complementary approaches to transforming agent execution experience into persistent, reusable capabilities. Experience- and memory-based methods accumulate workflows, reasoning strategies, and procedural knowledge from past interactions, enabling knowledge reuse across tasks \citep{wang2025awm,ouyang2026reasoningbank,fang2026memp}. Trajectory-to-skill approaches further distill successful and failed executions into reusable skill artifacts and executable routines \citep{zheng2025skillweaver,ni2026trace2skill,alzubi2026evoskill}. More recent methods focus on refining and managing evolving skills through evaluation feedback, failure analysis, behavioral validation, and lifecycle management \citep{yang2026skillopt,gautam2026skillaxe,moll2026grasp,zhang2026coevoskills,lin2026museautoskill}. Collectively, these studies have substantially advanced how agents acquire, refine, and maintain reusable capabilities from execution experience.

Despite these advances, existing approaches still largely organize learned knowledge around artifact boundaries, such as skill documents, memory entries, or skill banks, rather than systematically accounting for the distinct behavioral functions of individual updates. Yet a single execution trajectory can yield knowledge serving different purposes. For example, a spreadsheet task may suggest inspecting workbook structure before acting, verifying units before comparison, using a particular library to preserve formulas, or automating repeated sorting through a reusable script. Although derived from the same execution, these updates differ in scope, representation, validation requirements, and when they become useful. Organizing them around artifact boundaries alone does not determine how each should be maintained and used. The central challenge is therefore not merely what knowledge to retain, but how to organize, update, validate, and invoke it according to its functional role.

Cognitive science offers a useful perspective on how to organize such functionally diverse knowledge. Atkinson and Shiffrin distinguished task-dependent control processes from relatively stable memory structures \citep{atkinson1968human}. For agent skills, this distinction points to \emph{Meta} knowledge, which governs task-level planning, monitoring, verification, and stopping, while other knowledge supports specific decisions and actions. Yet the knowledge underlying these actions is itself diverse. Cohen and Squire's distinction between knowing how and knowing that \citep{cohen1980preserved} highlights differences in how knowledge can be learned and retained. In agent skills, \emph{Experience} captures context-conditioned heuristics and recovery strategies, \emph{Tool} provides capability-specific operational guidance, and \emph{Script} encodes reusable executable procedures with execution and validation contracts. These functional categories are tailored to agent execution rather than directly mapped from human cognitive systems. Finally, Miyake et al. highlighted the unity and diversity of executive functions \citep{miyake2000unity}, underscoring the importance of coordination among specialized functions. Together, these perspectives suggest that different forms of skill knowledge should be maintained according to their functional roles while remaining coordinated during task execution.

\begin{figure}[t]
  \centering
  \includegraphics[width=0.8\columnwidth]{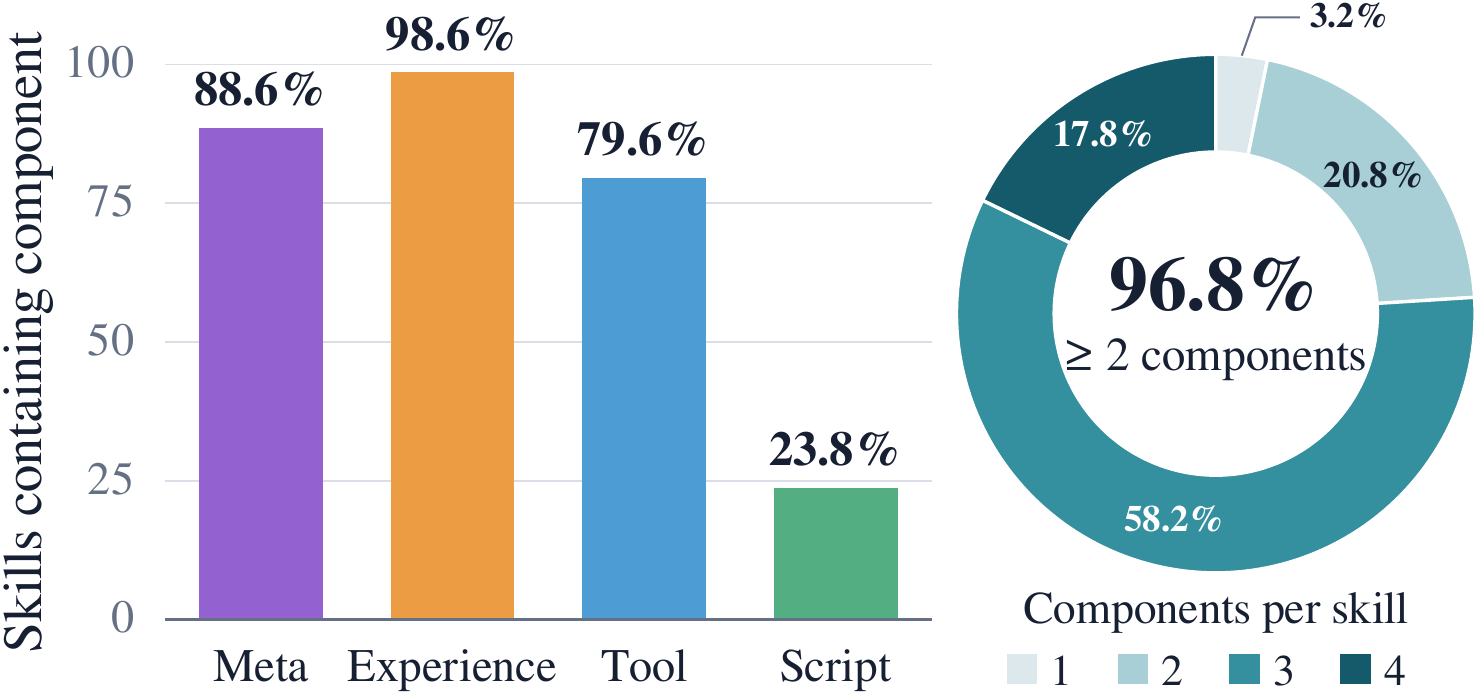}
  \caption{Functional composition of 500 widely used skill packages. Left:
  prevalence of each component; right: the number of components per skill.
  Component labels are non-exclusive.}
  \label{fig:intro-skill-statistics}
\end{figure}

To examine these functional distinctions in real-world agent skills, we analyze 500 widely used skill packages using an expert-derived annotation scheme (Appendix~\ref{app:skill-survey}). As shown in Figure~\ref{fig:intro-skill-statistics}, 96.8\% of the packages contain content spanning at least two functional roles, with Experience, Meta, and Tool appearing more frequently than Script. These findings suggest that multiple forms of reusable knowledge commonly coexist within individual skills, supporting a coordinated yet functionally differentiated approach to skill evolution.

Building on these cognitive insights and empirical observations, we propose \textbf{Cognitive Differentiation in Skill Evolution (CDSE)}, a framework that evolves a unified skill through four functionally distinct yet coordinated components. CDSE first performs \emph{Outcome-Contrastive Semantic Gradient Discovery} to extract evidence-grounded directions for reusable behavioral change from successful and failed trajectories. It then applies \emph{Cognition-Guided Gradient Consolidation} to identify which semantic gradients are worth retaining and assign them to Meta, Experience, Tool, or Script according to their functional roles. Finally, \emph{Adaptive Self-Improvement Update} refines these components through specialized update and validation mechanisms, while cross-component coordination and adaptive invocation support their use as task demands change. In this way, CDSE aligns how knowledge is learned from trajectories with how it is maintained and used during subsequent tasks.

Across six benchmarks, three agent models, and two execution harnesses, CDSE achieves the highest scores in all 36 evaluation settings, improving over the Base agent by 7.6 percentage points and over the strongest competing skill-evolution method in each setting by 2.6 points on average. Component-level analyses further reveal distinct patterns of contribution, invocation, and transfer. We summarize our main contributions as follows: (1) identifying and empirically characterizing functional heterogeneity in real-world agent skills; (2) proposing CDSE to organize and evolve trajectory-derived knowledge through function-specific yet coordinated mechanisms; and (3) systematically evaluating its effectiveness and examining the behavior of differentiated skill components across diverse tasks, models, and execution harnesses.

\section{Related Work}
\label{sec:related-work}

\paragraph{Language agents and self-improvement.}
LLM agents interleave reasoning, actions, and tool use to solve tasks through
environment interaction \citep{yao2023react,wang2024agentsurvey,yang2024sweagent}.
Self-improvement methods extend this loop by turning execution feedback into
persistent changes, either by retaining feedback, cross-task insights, and
reasoning strategies in context or memory
\citep{shinn2023reflexion,zhao2024expel,fu2024autoguide,
ouyang2026reasoningbank,zhang2026ace}, or by modifying the agent implementation
and even the meta-level adaptation procedure
\citep{zhang2026darwingodel,zhang2026hyperagents}.
Another line of work jointly evolves task generation and reasoning or
tool-using policies through reinforcement learning on self-generated
curricula \citep{zhao2025absolutezero,huang2026rzero,xia2026agent0}.
These works show that
improvement can persist at multiple levels of an agent system. CDSE focuses on
a complementary setting: the base model and improvement protocol remain fixed,
while evaluated trajectories update a persistent external skill state. The
central question is therefore how heterogeneous behavioral updates should be
organized within that evolving state.

\paragraph{Agent skills and skill evolution.}
Agent skills externalize reusable capabilities as instructions, workflows, or
executable procedures. Prior systems accumulate executable programs, reusable
action routines, or APIs from experience
\citep{wang2023voyager,wang2025awm,zheng2025skillweaver}. Agent Skills packages
broaden this representation to instructions, linked resources, and scripts
loaded through progressive disclosure \citep{zhang2025agentskills}, while
SkillsBench shows that skill efficacy depends on the task, model, harness, and
package design \citep{li2026skillsbench}. Recent trajectory-driven methods
create and refine skill artifacts from successful and failed executions using
failure analysis, bounded edits, and evaluation feedback
\citep{ni2026trace2skill,alzubi2026evoskill,yang2026skillopt,gautam2026skillaxe},
with subsequent work improving validation, lifecycle management, memory,
multi-file evolution, and skill or meta-skill adaptation
\citep{moll2026grasp,he2026skillcommit,zhang2026librarydrift,
zhang2026coevoskills,yang2026skillmaster,lin2026museautoskill,
shen2026skillalpha,wang2026metaskillevolve}. Despite these advances, the primary
unit of organization remains an artifact boundary---a skill document, folder,
or bank entry. Even when the artifact is hierarchical, multi-file, or
memory-enhanced, an update's execution function does not explicitly determine
its ownership, representation, update rule, validation contract, or invocation
timing. CDSE instead makes functional ownership the organizing principle,
assigning trajectory-derived updates to Meta, Experience, Tool, or Script and
maintaining them through differentiated but coordinated mechanisms.
\begin{figure*}[t]
  \centering
  \includegraphics[width=\textwidth]{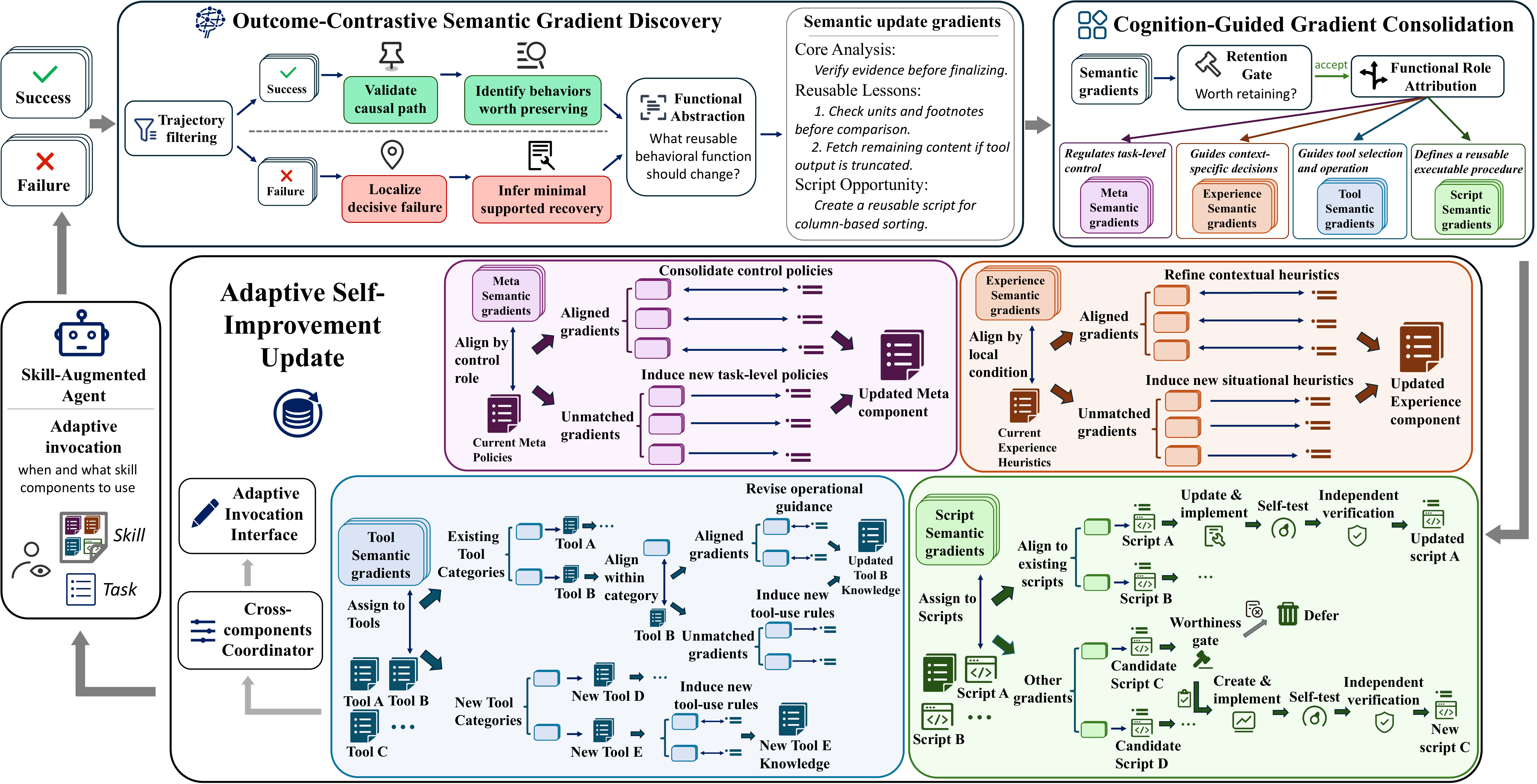}
  \caption{CDSE discovers semantic gradients from successful and failed
  executions, consolidates them by functional role, and updates the four
  components through specialized maintenance and coordination. Revised
  invocation descriptions connect the evolving skill to subsequent agent
  behavior.}
  \label{fig:method-overview}
\end{figure*}

\section{Cognitive Differentiation in Skill Evolution}
\label{sec:method}

CDSE evolves a unified skill by discovering semantic gradients from evaluated
trajectories, consolidating them by functional role, and updating its
components through specialized maintenance and cross-component coordination
(Figure~\ref{fig:method-overview}). The updated components are then invoked
adaptively during subsequent tasks.

\subsection{Problem Formulation: Trajectory-Driven Skill Self-Evolution}
\label{sec:method-formulation}

An agent $\pi_\theta$ solves a task $x$ by interacting with an
environment while selectively loading relevant components of a
persistent skill. At evolution round $t$, the unified skill state
comprises four functionally differentiated components: Meta, Experience, Tool, and Script. We denote this state by
$\mathcal{B}^{(t)}=(\mathcal{M},\mathcal{E},\mathcal{T},
\mathcal{S})^{(t)}$, with component indices
$\mathcal{K}=\{M,E,T,S\}$.

Let $\mathcal{D}^{(t)}=\{(\tau_i,y_i)\}_{i=1}^{N_t}$
denote the evaluated trajectory corpus, where $\tau_i$ is
an execution trace and $y_i$ its evaluated outcome.
CDSE aims to improve subsequent task performance by updating
the persistent skill state as follows:
\begin{equation}
  \mathcal{B}^{(t+1)}=
  \mathcal{U}\!\left(\mathcal{B}^{(t)},
  \mathcal{R}\big(\mathcal{G}(\mathcal{D}^{(t)})\big)\right),
  \label{eq:cdse-loop}
\end{equation}
where $\mathcal{G}$ extracts candidate semantic gradients,
$\mathcal{R}$ retains and assigns them to functional components,
and $\mathcal{U}$ performs the validated skill update detailed
in Section~\ref{sec:typed-evolution}.
This process updates only the external skill state, leaving
the model parameters and update protocol fixed. The formulation
also supports repeated consolidation over a fixed trajectory
corpus, as examined in our multi-round experiments.

\subsection{Outcome-Contrastive Semantic Gradient Discovery}
\label{sec:finding-induction}

The discovery stage extracts candidate semantic gradients from evaluated
trajectories, i.e., trace-grounded, reusable directions for behavioral
change \citep{shinn2023reflexion,pryzant2023automatic}.
A deterministic filter $P$ removes runtime-only artifacts while preserving
reasoning, tool actions, and observations relevant to behavioral diagnosis.
Successful and failed trajectories are then analyzed under distinct
diagnostic objectives. For successful trajectories, the analyzer identifies
trace-supported behaviors worth preserving and safeguards for fragile
successes. For failed trajectories, it localizes evidenced decisive errors
or missing transitions and identifies the minimal supported recovery.
An observed outcome alone does not establish the validity of a procedure
or diagnosis; unsupported causes remain unresolved. 

Let $s(y_i)\in\{+,-\}$ denote the success/failure label derived from
the evaluated outcome $y_i$. The corresponding analyzer $A_{s(y_i)}$
processes the filtered trajectory and its outcome to produce candidate
semantic gradients:
\begin{equation}
  \mathcal{F}=\mathcal{G}(\mathcal{D}^{(t)})=\bigcup_{i=1}^{N_t}A_{s(y_i)}(P(\tau_i),y_i).
  \label{eq:gradient-discovery}
\end{equation}
Each $f=(g,e)\in\mathcal{F}$ pairs a proposed reusable behavioral update
$g$ with its supporting trace lines $e$. The analyzer produces a core
analysis, reusable lessons, and, when applicable, a script opportunity.
Supporting evidence remains attached to each candidate, providing the
basis for subsequent retention and functional attribution.

\subsection{Cognition-Guided Gradient Consolidation}
\label{sec:gradient-consolidation}

Drawing on cognitive distinctions between control processes and memory
structures, as well as between forms of knowing
\citep{atkinson1968human,cohen1980preserved}, CDSE separates gradient
consolidation into two steps: retention and functional attribution.
Retention determines which candidate gradients are worth preserving,
while attribution identifies the components responsible for maintaining
them.

A retention gate $r(f)\in\{0,1\}$ selects evidence-supported, actionable,
and reusable gradients, filtering out generic, instance-specific,
unsupported, or redundant findings. Each retained gradient is assigned
one primary component and, when its evidence supports an additional
functional role, at most one secondary component.
Let $\rho(f)\subseteq\mathcal{K}$ denote the set of assigned components,
with $\rho(f)=\varnothing$ if $r(f)=0$. Consolidation produces
\begin{equation}
  \begin{aligned}
  \mathcal{F}_k&=\{f\in\mathcal{F}:r(f)=1,\ k\in\rho(f)\},\\
  k&\in\mathcal{K},\quad
  1\leq|\rho(f)|\leq2\ \text{if }r(f)=1.
  \end{aligned}
  \label{eq:routing}
\end{equation}
Functional attribution follows the behavioral role of each gradient.
Meta maintains task-level control policies, including planning,
monitoring, and global verification. Experience maintains
context-conditioned heuristics and recovery strategies activated by
local conditions. Tool maintains capability-specific operational
guidance, while Script receives proposals for reusable executable
procedures and their execution or validation contracts. The primary
assignment is recorded separately, and each gradient retains its
original content and supporting evidence for subsequent component-specific updates.

\subsection{Adaptive Self-Improvement Update}
\label{sec:typed-evolution}

Given the consolidated gradient sets
$\{\mathcal{F}_k\}_{k\in\mathcal{K}}$, the update
operator $\mathcal{U}$ in Equation~\ref{eq:cdse-loop}
combines component-specific maintenance operators $U_k$,
cross-component coordination $C$, and
invocation-description rewriting $D$:
\begin{equation}
  \begin{aligned}
  \widehat{\mathcal{B}}_k^{(t+1)}
  &=U_k(\mathcal{B}_k^{(t)},\mathcal{F}_k),\\
  \mathcal{B}^{(t+1)}
  &=D\!\left(C(\widehat{\mathcal{B}}^{(t+1)})\right).
  \end{aligned}
  \label{eq:type-update}
\end{equation}
Here, $\widehat{\mathcal{B}}_k^{(t+1)}$ denotes
the candidate update to component $k$, and
$\widehat{\mathcal{B}}^{(t+1)}$ collects the four
component updates. Inspired by the unity and diversity
of cognitive functions \citep{miyake2000unity},
this design combines specialized maintenance
with coordination across the unified skill state.

\paragraph{Specialized updates.}
Meta and Experience maintain atomic behavioral rules.
Both align incoming gradients with existing rules,
aggregate supporting evidence for revisions, and
induce new rules from unmatched gradients.
Meta updates task-level control policies, whereas
Experience organizes heuristics around explicit
local activation conditions.
Tool assigns gradients to capability-specific
categories and revises or induces operational rules
within each category. Category-level capacity control
preserves coverage across different tools.
Script aligns proposals with existing procedures
and screens new proposals for deterministic reuse,
explicit input/output contracts, and feasible tests.
Accepted proposals are implemented or revised,
self-tested, and independently verified.
Unverified executable changes are withheld.

\paragraph{Cross-component coordination.}
\label{sec:coordination}
After specialized maintenance, the coordinator $C$
resolves conflicting or redundant guidance and
residual ownership ambiguity across components
through minimal edits that preserve verified
executable artifacts and component-specific
responsibilities.

\paragraph{Adaptive invocation.}
The description rewriter $D$ derives each component's
scope, applicable conditions, and expected contribution
from the coordinated skill state.
Visible before component loading, these descriptions
guide the agent's selection of relevant components
through its ordinary invocation mechanism as task
demands change.
The updated skill state is published only after
coordination and description changes pass their
respective validation checks.
Details are in Appendix~\ref{app:method-details}.

\begin{table*}[!t]
  \centering
  \newcommand{\basecell}[1]{\makebox[20pt][r]{#1}\makebox[20pt][l]{}}
  \newcommand{\scoregain}[2]{\makebox[20pt][r]{#1}\makebox[20pt][l]{\hspace{0.06em}{\color{gaincolor}\fontsize{8.0}{8.5}\selectfont #2}}}
  \newcommand{\bestgain}[2]{\makebox[20pt][r]{\textbf{#1}}\makebox[20pt][l]{\hspace{0.06em}{\color{gaincolor}\fontsize{8.0}{8.5}\selectfont\bfseries #2}}}
  \newcommand{\secondgain}[2]{\makebox[20pt][r]{\underline{#1}}\makebox[20pt][l]{\hspace{0.06em}{\color{gaincolor}\fontsize{8.0}{8.5}\selectfont #2}}}
  \normalsize
  \setlength{\tabcolsep}{0.8pt}
  \renewcommand{\arraystretch}{1.02}
  \begin{tabularx}{\textwidth}{>{\small}l>{\small}l*{6}{>{\centering\arraybackslash}X}}
    \toprule
    {\small Agent model} & {\small Method} & {\small OfficeQA} & {\small\makebox[0pt][c]{Spreadsheet2}} & {\small SealQA} & {\small LiveMath} & {\small SWE-Pro} & {\small Automation} \\
    \midrule
    \multirow{5}{*}{\shortstack[l]{GLM-5.2\\(Claude Code)}}
      & Base          & \basecell{71.5} & \basecell{24.0} & \basecell{45.9} & \basecell{83.0} & \basecell{60.2} & \basecell{20.8} \\
      & Trace2Skill   & \scoregain{74.1}{+2.6} & \secondgain{28.3}{+4.3} & \scoregain{48.2}{+2.3} & \scoregain{82.3}{-0.7} & \scoregain{62.6}{+2.4} & \scoregain{21.4}{+0.6} \\
      & SkillOpt-lite & \secondgain{76.4}{+4.9} & \scoregain{26.8}{+2.8} & \scoregain{49.7}{+3.8} & \secondgain{85.0}{+2.0} & \secondgain{63.1}{+2.9} & \scoregain{24.7}{+3.9} \\
      & EvoSkill      & \scoregain{76.0}{+4.5} & \scoregain{25.9}{+1.9} & \secondgain{52.4}{+6.5} & \scoregain{84.4}{+1.4} & \scoregain{62.6}{+2.4} & \secondgain{25.5}{+4.7} \\
    \rowcolor{oursgray}
      \cellcolor{white} & \textbf{CDSE} & \bestgain{77.2}{+5.7} & \bestgain{30.7}{+6.7} & \bestgain{54.1}{+8.2} & \bestgain{87.7}{+4.7} & \bestgain{68.4}{+8.2} & \bestgain{32.9}{+12.1} \\
    \midrule
    \multirow{5}{*}{\shortstack[l]{DeepSeek-V4-Flash\\(Claude Code)}}
      & Base          & \basecell{68.0} & \basecell{16.8} & \basecell{51.8} & \basecell{59.9} & \basecell{57.8} & \basecell{14.3} \\
      & Trace2Skill   & \scoregain{70.1}{+2.1} & \secondgain{24.2}{+7.4} & \scoregain{54.3}{+2.5} & \scoregain{62.3}{+2.4} & \scoregain{56.8}{-1.0} & \scoregain{19.2}{+4.9} \\
      & SkillOpt-lite & \secondgain{75.4}{+7.4} & \scoregain{22.7}{+5.9} & \scoregain{55.9}{+4.1} & \scoregain{63.9}{+4.0} & \secondgain{60.0}{+2.2} & \secondgain{22.6}{+8.3} \\
      & EvoSkill      & \scoregain{72.8}{+4.8} & \secondgain{24.2}{+7.4} & \secondgain{56.5}{+4.7} & \secondgain{65.7}{+5.8} & \scoregain{59.6}{+1.8} & \scoregain{21.3}{+7.0} \\
    \rowcolor{oursgray}
      \cellcolor{white} & \textbf{CDSE} & \bestgain{77.3}{+9.3} & \bestgain{26.9}{+10.1} & \bestgain{59.4}{+7.6} & \bestgain{69.1}{+9.2} & \bestgain{64.1}{+6.3} & \bestgain{28.9}{+14.6} \\
    \midrule
    \multirow{5}{*}{\shortstack[l]{MiniMax-M3\\(Claude Code)}}
      & Base          & \basecell{71.6} & \basecell{25.3} & \basecell{49.4} & \basecell{73.5} & \basecell{57.2} & \basecell{9.1} \\
      & Trace2Skill   & \scoregain{73.2}{+1.6} & \secondgain{33.6}{+8.3} & \scoregain{53.1}{+3.7} & \scoregain{77.6}{+4.1} & \scoregain{62.0}{+4.8} & \scoregain{13.3}{+4.2} \\
      & SkillOpt-lite & \scoregain{75.1}{+3.5} & \scoregain{31.9}{+6.6} & \secondgain{54.8}{+5.4} & \scoregain{79.5}{+6.0} & \scoregain{61.2}{+4.0} & \scoregain{15.0}{+5.9} \\
      & EvoSkill      & \secondgain{76.3}{+4.7} & \scoregain{30.5}{+5.2} & \scoregain{54.2}{+4.8} & \secondgain{80.1}{+6.6} & \secondgain{62.9}{+5.7} & \secondgain{15.5}{+6.4} \\
    \rowcolor{oursgray}
      \cellcolor{white} & \textbf{CDSE} & \bestgain{81.8}{+10.2} & \bestgain{34.4}{+9.1} & \bestgain{57.9}{+8.5} & \bestgain{82.7}{+9.2} & \bestgain{65.3}{+8.1} & \bestgain{20.0}{+10.9} \\
    \midrule
    \multirow{5}{*}{\shortstack[l]{GLM-5.2\\(Codex)}}
      & Base          & \basecell{68.8} & \basecell{22.7} & \basecell{22.6} & \basecell{79.0} & \basecell{62.3} & \basecell{42.2} \\
      & Trace2Skill   & \scoregain{70.8}{+2.0} & \secondgain{29.8}{+7.1} & \scoregain{24.1}{+1.5} & \scoregain{80.5}{+1.5} & \scoregain{64.8}{+2.5} & \scoregain{43.8}{+1.6} \\
      & SkillOpt-lite & \secondgain{72.9}{+4.1} & \scoregain{28.5}{+5.8} & \secondgain{25.9}{+3.3} & \scoregain{81.1}{+2.1} & \scoregain{65.7}{+3.4} & \scoregain{44.3}{+2.1} \\
      & EvoSkill      & \scoregain{71.6}{+2.8} & \scoregain{29.4}{+6.7} & \scoregain{24.0}{+1.4} & \secondgain{81.5}{+2.5} & \secondgain{67.0}{+4.7} & \secondgain{44.9}{+2.7} \\
    \rowcolor{oursgray}
      \cellcolor{white} & \textbf{CDSE} & \bestgain{74.9}{+6.1} & \bestgain{31.3}{+8.6} & \bestgain{27.6}{+5.0} & \bestgain{83.5}{+4.5} & \bestgain{68.9}{+6.6} & \bestgain{47.1}{+4.9} \\
    \midrule
    \multirow{5}{*}{\shortstack[l]{DeepSeek-V4-Flash\\(Codex)}}
      & Base          & \basecell{66.7} & \basecell{30.0} & \basecell{36.5} & \basecell{64.2} & \basecell{58.9} & \basecell{37.0} \\
      & Trace2Skill   & \scoregain{68.1}{+1.4} & \secondgain{36.2}{+6.2} & \scoregain{37.4}{+0.9} & \scoregain{65.9}{+1.7} & \scoregain{61.5}{+2.6} & \scoregain{40.5}{+3.5} \\
      & SkillOpt-lite & \scoregain{68.9}{+2.2} & \scoregain{34.3}{+4.3} & \scoregain{38.0}{+1.5} & \scoregain{66.5}{+2.3} & \scoregain{62.9}{+4.0} & \scoregain{41.7}{+4.7} \\
      & EvoSkill      & \secondgain{69.8}{+3.1} & \scoregain{35.4}{+5.4} & \secondgain{38.2}{+1.7} & \secondgain{67.5}{+3.3} & \secondgain{63.6}{+4.7} & \secondgain{43.1}{+6.1} \\
    \rowcolor{oursgray}
      \cellcolor{white} & \textbf{CDSE} & \bestgain{71.9}{+5.2} & \bestgain{38.1}{+8.1} & \bestgain{40.8}{+4.3} & \bestgain{69.2}{+5.0} & \bestgain{65.9}{+7.0} & \bestgain{45.4}{+8.4} \\
    \midrule
    \multirow{5}{*}{\shortstack[l]{MiniMax-M3\\(Codex)}}
      & Base          & \basecell{65.1} & \basecell{18.7} & \basecell{21.5} & \basecell{71.0} & \basecell{54.1} & \basecell{34.4} \\
      & Trace2Skill   & \scoregain{67.5}{+2.4} & \scoregain{27.2}{+8.5} & \scoregain{25.0}{+3.5} & \scoregain{72.9}{+1.9} & \scoregain{56.7}{+2.6} & \scoregain{36.3}{+1.9} \\
      & SkillOpt-lite & \secondgain{70.9}{+5.8} & \secondgain{29.1}{+10.4} & \scoregain{25.3}{+3.8} & \secondgain{73.5}{+2.5} & \scoregain{57.1}{+3.0} & \secondgain{38.7}{+4.3} \\
      & EvoSkill      & \scoregain{68.6}{+3.5} & \scoregain{25.3}{+6.6} & \secondgain{26.2}{+4.7} & \scoregain{72.2}{+1.2} & \secondgain{57.5}{+3.4} & \scoregain{37.1}{+2.7} \\
    \rowcolor{oursgray}
      \cellcolor{white} & \textbf{CDSE} & \bestgain{72.2}{+7.1} & \bestgain{29.4}{+10.7} & \bestgain{29.4}{+7.9} & \bestgain{74.5}{+3.5} & \bestgain{60.8}{+6.7} & \bestgain{39.9}{+5.5} \\
    \bottomrule
  \end{tabularx}
  \caption{Task scores (\%) across models, harnesses, and skill-evolution methods.}
  \label{tab:main-results}
\end{table*}

\section{Experiments}
\label{sec:experiments}

\subsection{Experimental Setup}
\label{sec:experimental-setup}

We evaluate on six benchmarks: OfficeQA~\citep{opsahlong2026officeqa},
SpreadsheetBench2~\citep{zhu2026spreadsheetbench2},
SealQA~\citep{pham2026sealqa}, LiveMathBench~\citep{liu2025livemathbench},
SWE-Pro~\citep{deng2025swebenchpro}, and
AutomationBench~\citep{shepard2026automationbench}, using
GLM-5.2~\citep{glm5team2026glm5},
DeepSeek-V4-Flash~\citep{deepseekai2026deepseekv4}, and
MiniMax-M3~\citep{lai2026minimaxsparseattention} under both Claude Code and
Codex. Each
benchmark collection is split into training and test sets at a 2:7 ratio.
Training trajectories support skill evolution; all reported results use the
test split and benchmark-specific scoring procedures.

We compare CDSE with the same agent without an evolved skill state (Base),
Trace2Skill~\citep{ni2026trace2skill}, SkillOpt-lite~\citep{yang2026skillopt},
and EvoSkill~\citep{alzubi2026evoskill}. Within each model--harness block,
methods share the training split, task instances, tool access, and context
limit. Table~\ref{tab:main-results} reports results after three evolution
rounds for each skill-evolution method. Task execution is bounded by
benchmark-specific per-question time and turn limits; tasks that remain
unfinished when either limit is reached are counted as failures.
Aggregate improvements are macro-averages of within-benchmark
percentage-point changes over the relevant model--harness--benchmark cells.
Dataset selections, the evaluation protocol, and baseline configurations are
detailed in Appendix~\ref{app:experimental-details}.

\begin{figure*}[!t]
  \centering
  \includegraphics[width=0.95\textwidth]{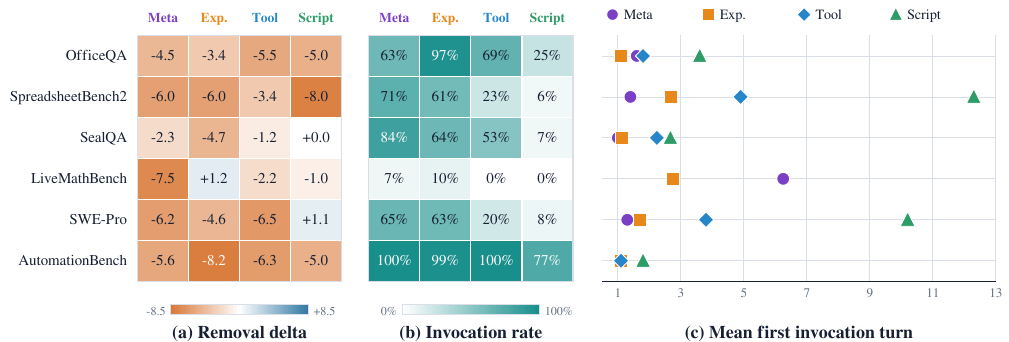}
  \caption{Component removal effects and invocation behavior on GLM-5.2.}
  \label{fig:skill-specialization}
\end{figure*}

\subsection{Overall Performance}
\label{sec:main-results}

Table~\ref{tab:main-results} shows that CDSE achieves the highest score in
all 36 settings across six benchmarks, three agent models, and two execution
harnesses. On average, it improves over Base by 7.6 percentage points and
exceeds the strongest competing skill-evolution baseline in each setting by
2.6 points. This advantage holds on every benchmark with both Claude Code
and Codex, spanning document and mathematical reasoning, spreadsheet
manipulation, software engineering, and business workflows. These results support CDSE across diverse tasks and execution environments. With GLM-5.2 and Claude Code, single-round token costs are
comparable across methods (Appendix~\ref{app:token-consumption}).

\subsection{Functional Differentiation and Adaptive Invocation}
\label{sec:skill-analysis}

Figure~\ref{fig:skill-specialization} examines component-removal effects (a),
invocation rates (b), and first-invocation timing (c). Invocation patterns
vary across benchmarks: all four components are invoked in at
least 76.7\% of AutomationBench episodes, whereas Script is invoked in only
6.0\% on SpreadsheetBench2. Invocation timing also differs, with Script
loaded later on SpreadsheetBench2 and SWE-Pro, but Tool and Script loaded
early on AutomationBench.

Component contributions are similarly task-dependent. Removing Tool
reduces performance by 1.2 points on SealQA but 6.5 on SWE-Pro, while
removing any component on AutomationBench causes a 5.0--8.2-point drop.
Notably, despite its low invocation rate, removing Script causes the
largest drop (8.0 points) on SpreadsheetBench2. Component importance therefore does not directly follow invocation frequency.
Under progressive disclosure, descriptions remain visible before loading,
so even components with zero recorded invocations can influence execution
and exhibit ablation effects. Appendix~\ref{app:component-analysis} reports
mean invocation turns and further component analysis.

\begin{figure}[htbp]
  \centering
  \includegraphics[width=\columnwidth]{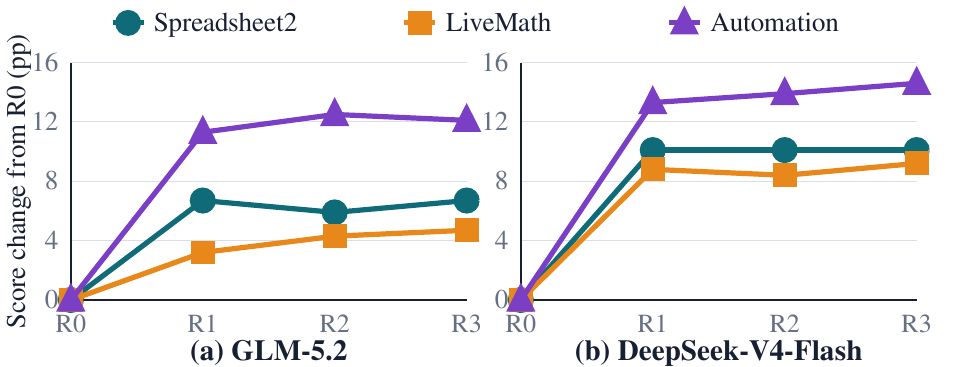}
  \caption{Score gains across three CDSE evolution rounds.}
  \label{fig:evolution-rounds}
\end{figure}

\subsection{Multi-Round Skill Evolution}
\label{sec:evolution-rounds}

Figure~\ref{fig:evolution-rounds} compares three CDSE evolution rounds over
the same trajectory corpus on SpreadsheetBench2, LiveMathBench, and
AutomationBench, using GLM-5.2 and DeepSeek-V4-Flash under Claude Code.
The first round delivers most of the improvement; later rounds show smaller,
benchmark-dependent changes and non-monotonic fluctuations. Results across
all six benchmarks and three models show the same diminishing-return
pattern under this fixed-corpus setting (Appendix~\ref{app:evolution-results}).

\begin{figure}[htbp]
  \centering
  \includegraphics[width=\columnwidth]{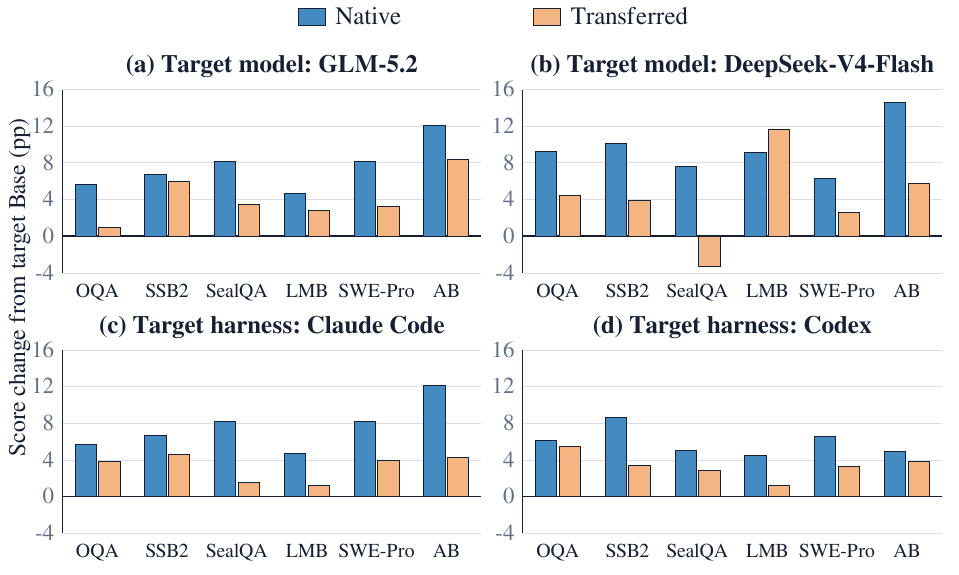}
  \caption{Cross-model and cross-harness transfer gains.}
  \label{fig:cross-model-harness-transfer}
\end{figure}

\subsection{Transfer across Tasks, Models, and Harnesses}
\label{sec:transfer}

\begin{table*}[!t]
  \centering
  \captionsetup{font=small}
  \newcommand{\transferpos}[2]{{\color{gaincolor}\bfseries#1}\hspace{0.10em}{\color{samecolor}\fontsize{7.0}{7.5}\selectfont\bfseries(#2\%)}}
  \newcommand{\transferneg}[2]{{\color{losscolor}\bfseries#1}\hspace{0.10em}{\color{samecolor}\fontsize{7.0}{7.5}\selectfont\bfseries(#2\%)}}
  \newcommand{\transferzero}[2]{{\color{samecolor}\bfseries#1}\hspace{0.10em}{\color{samecolor}\fontsize{7.0}{7.5}\selectfont\bfseries(#2\%)}}
  \small
  \setlength{\tabcolsep}{1.2pt}
  \renewcommand{\arraystretch}{1.10}
  \begin{tabular*}{0.9\textwidth}{@{\extracolsep{\fill}}>{\small}l>{\small}lcccccc@{}}
    \toprule
    {\small Source} & {\small Component} & {\small OfficeQA} & {\small Spreadsheet2} & {\small SealQA} & {\small LiveMath} & {\small SWE-Pro} & {\small Automation} \\
    \midrule
    \multirow{4}{*}{Spreadsheet2}
      & Meta       & \transferpos{+4.1}{18} & -- & \transferpos{+4.3}{20} & \transferzero{+0.0}{0} & \transferpos{+4.1}{10} & \transferpos{+7.1}{28} \\
      & Experience & \transferpos{+0.8}{6}  & -- & \transferneg{-2.8}{14} & \transferneg{-0.4}{0} & \transferzero{+0.0}{7}  & \transferneg{-1.6}{2} \\
      & Tool       & \transferneg{-1.3}{1}  & -- & \transferzero{+0.0}{0}  & \transferpos{+1.0}{0} & \transferneg{-0.6}{1}  & \transferzero{+0.0}{0} \\
      & Script     & \transferneg{-0.6}{2}  & -- & \transferneg{-0.9}{0}  & \transferpos{+0.7}{0} & \transferneg{-3.9}{3}  & \transferneg{-1.6}{0} \\
    \midrule
    \multirow{4}{*}{Automation}
      & Meta       & \transferpos{+5.7}{10} & \transferpos{+1.4}{19} & \transferpos{+6.7}{35} & \transferpos{+0.2}{0} & \transferpos{+2.6}{6}  & -- \\
      & Experience & \transferpos{+2.2}{7}  & \transferneg{-1.2}{0}  & \transferpos{+3.8}{38} & \transferneg{-0.8}{0} & \transferzero{+0.0}{4}  & -- \\
      & Tool       & \transferneg{-1.6}{0}  & \transferneg{-0.5}{0}  & \transferneg{-1.1}{0}  & \transferzero{+0.0}{0} & \transferneg{-0.6}{0}  & -- \\
      & Script     & \transferzero{+0.0}{0}  & \transferzero{+0.0}{0}  & \transferzero{+0.0}{0}  & \transferneg{-0.4}{0} & \transferzero{+0.0}{0}  & -- \\
    \bottomrule
  \end{tabular*}
  \caption{Cross-benchmark component transfer on GLM-5.2.}
  \label{tab:cross-benchmark-transfer}
\end{table*}

\paragraph{Cross-model and cross-harness transfer.}
Figure~\ref{fig:cross-model-harness-transfer} evaluates the transfer of
complete skill states between GLM-5.2 and DeepSeek-V4-Flash, and between
Claude Code and Codex. Cross-model transfer improves 11 of 12 target
configurations, averaging 4.2 percentage points (pp) over Base for each
target model. Cross-harness transfer improves all 12 configurations,
averaging 3.3 pp for Claude Code and 3.4 pp for Codex. Although native
skill states perform better on average, transferred states retain
substantial gains across models and harnesses
(Appendix~\ref{app:transfer-details}).

\paragraph{Cross-benchmark component transfer.}
Table~\ref{tab:cross-benchmark-transfer} evaluates component
transfer from SpreadsheetBench2 or AutomationBench to other benchmarks.
Meta improves 9 of 10 source--target pairs, averaging 3.6 pp over Base,
whereas Experience transfers more selectively and Tool and Script show
limited consistent gains. These differences are consistent with the
broader applicability of task-level control knowledge and the stronger
context or environment dependence of other components. Parenthesized
values report explicit component-loading rates; descriptions remain
visible before loading, so zero loading does not imply zero influence.

\subsection{Mechanism Ablations}
\label{sec:mechanism-ablations}

\begin{table}[H]
  \centering
  \footnotesize
  \setlength{\tabcolsep}{3pt}
  \renewcommand{\tabularxcolumn}[1]{m{#1}}
  \begin{tabularx}{\columnwidth}{@{}>{\raggedright\arraybackslash}Xccc@{}}
    \toprule
    Variant & \shortstack{Spreadsheet\\Bench2} & SWE-Pro & \shortstack{Automation\\Bench} \\
    \midrule
    Base & 24.0 & 60.2 & 20.8 \\
    \arrayrulecolor{black!30}\midrule[0.25pt]\arrayrulecolor{black}
    Full CDSE & \textbf{30.7} & \textbf{68.4} & \textbf{32.9} \\
    \arrayrulecolor{black!30}\midrule[0.25pt]\arrayrulecolor{black}
    w/o Cognitive Differentiation & 25.8 & 62.6 & 24.3 \\
    \arrayrulecolor{black!30}\midrule[0.25pt]\arrayrulecolor{black}
    w/o Specialized Update & 27.1 & 65.4 & 28.0 \\
    \arrayrulecolor{black!30}\midrule[0.25pt]\arrayrulecolor{black}
    w/o Cross-Component Coordination & 29.6 & 67.5 & 30.7 \\
    \arrayrulecolor{black!30}\midrule[0.25pt]\arrayrulecolor{black}
    w/o Adaptive Invocation Interface & 26.4 & 64.3 & 27.1 \\
    \bottomrule
  \end{tabularx}
  \caption{Mechanism ablations on GLM-5.2 with Claude Code. All entries are task scores (\%).}
  \label{tab:mechanism-ablations}
\end{table}

Table~\ref{tab:mechanism-ablations} evaluates four mechanism ablations on three benchmarks. Removing cognitive
differentiation collapses the four components into a unified skill,
whereas removing specialized updates preserves the component structure
but applies a shared update strategy. The other variants disable
cross-component coordination or remove the Adaptive Invocation Interface
that provides content-aware descriptions to guide component selection.

Full CDSE achieves the highest score across all three benchmarks.
Removing cognitive differentiation causes the largest drops,
followed by removing the Adaptive Invocation Interface.
The no-differentiation variant uses the same semantic gradients as CDSE, suggesting benefits from functional organization and maintenance beyond lesson extraction.
Specialized updates and cross-component coordination also contribute.
These results highlight the importance of functional organization,
specialized maintenance, and description-guided component selection.
Appendix~\ref{app:mechanism-ablations} details the four variants.

\subsection{Token Efficiency}
\label{sec:token-efficiency}

Using GLM-5.2 with Claude Code, we divide each method's single-round
input-token count by its score gain over Base in the same round.
Table~\ref{tab:token-efficiency} reports the resulting cost per percentage
point of improvement. CDSE achieves the lowest cost on all three benchmarks
with comparable input-token consumption (Appendix~\ref{app:token-consumption}),
indicating improved token efficiency without substantially increasing
the evolution budget.

\begin{table}[H]
  \centering
  \small
  \setlength{\tabcolsep}{3pt}
  \begin{tabularx}{\columnwidth}{@{}l*{3}{>{\centering\arraybackslash}X}@{}}
    \toprule
    Method & \shortstack{Spreadsheet2} & SWE-Pro & \shortstack{Automation} \\
    \midrule
    Trace2Skill & 15.41 & 41.30 & 23.90 \\
    SkillOpt-lite & 29.33 & 59.82 & 6.56 \\
    EvoSkill & 42.67 & 75.47 & 5.06 \\
    \rowcolor{oursgray}
    \textbf{CDSE} & \textbf{7.60} & \textbf{15.04} & \textbf{2.19} \\
    \bottomrule
  \end{tabularx}
  \caption{Input tokens per percentage point of score gain over Base after
  one evolution round (million tokens/pp, $\downarrow$), using GLM-5.2
  with Claude Code.}
  \label{tab:token-efficiency}
\end{table}

\FloatBarrier
\section{Conclusion}
CDSE organizes trajectory-derived knowledge into four functionally distinct components—Meta, Experience, Tool, and Script—using specialized updates, cross-component coordination, and adaptive invocation. Across six benchmarks, three models, and two harnesses, it achieves the highest scores in all 36 settings. Component and transfer analyses reveal task-dependent contributions and varying transferability, while ablations highlight the importance of functional differentiation and adaptive invocation. Token-efficiency analysis further shows that CDSE achieves greater performance gains per input token. These findings suggest that effective skill evolution benefits from maintaining and using knowledge according to its functional role.

\section*{Limitations}
Our evaluation covers six benchmarks, three agent models, and two execution
harnesses, leaving generalization to other domains and agent configurations
to future work.
Multi-round experiments reuse a fixed trajectory corpus and therefore do
not establish sustained improvement in continually changing environments.
The four-component organization is supported by our survey and experiments,
but alternative decompositions and their effectiveness remain to be studied.

\section*{Ethical Considerations}
CDSE is intended to support research on reliable and reusable LLM agents.
Evolved skills may retain errors or biases from models and execution
trajectories, so improved task performance does not guarantee safe deployment.
Practical use should respect data privacy and third-party licenses, restrict
tool permissions, and retain human oversight for consequential actions.
We encourage responsible use of the framework and derived artifacts.

\bibliography{references}

\FloatBarrier
\appendix
\makeatletter
\setlength{\@fptop}{0pt}
\setlength{\@fpsep}{12pt}
\setlength{\@fpbot}{0pt plus 1fil}
\setlength{\@dblfptop}{0pt}
\setlength{\@dblfpsep}{12pt}
\setlength{\@dblfpbot}{0pt plus 1fil}
\makeatother
\renewcommand{\topfraction}{0.95}
\renewcommand{\dbltopfraction}{0.95}
\renewcommand{\textfraction}{0.05}
\renewcommand{\floatpagefraction}{0.75}
\renewcommand{\dblfloatpagefraction}{0.75}
\setcounter{topnumber}{5}
\setcounter{totalnumber}{8}
\setcounter{dbltopnumber}{4}
\setlength{\textfloatsep}{10pt plus 2pt minus 2pt}
\setlength{\dbltextfloatsep}{10pt plus 2pt minus 2pt}
\setlength{\floatsep}{10pt plus 2pt minus 2pt}
\setlength{\dblfloatsep}{10pt plus 2pt minus 2pt}
\raggedbottom

\section{Token Consumption and Efficiency for One Evolution Round}
\label{app:token-consumption}

Table~\ref{tab:single-round-tokens} reports the input and output token counts
for one evolution round with GLM-5.2 and Claude Code.

\begin{table*}[!b]
  \centering
  \footnotesize
  \setlength{\tabcolsep}{3pt}
  \begin{tabularx}{\textwidth}{@{}l*{12}{>{\centering\arraybackslash}X}@{}}
    \toprule
    & \multicolumn{2}{c}{OfficeQA}
    & \multicolumn{2}{c}{\shortstack{Spreadsheet\\Bench2}}
    & \multicolumn{2}{c}{SealQA}
    & \multicolumn{2}{c}{LiveMathBench}
    & \multicolumn{2}{c}{SWE-Pro}
    & \multicolumn{2}{c}{\shortstack{Automation\\Bench}} \\
    \cmidrule(lr){2-3}\cmidrule(lr){4-5}\cmidrule(lr){6-7}
    \cmidrule(lr){8-9}\cmidrule(lr){10-11}\cmidrule(lr){12-13}
    Method & Input & Output & Input & Output & Input & Output
    & Input & Output & Input & Output & Input & Output \\
    \midrule
    Trace2Skill & 35.80 & 1.350 & 49.30 & 2.700 & 10.60 & 0.480 & 9.51 & 1.050 & 123.90 & 5.600 & 23.90 & 1.050 \\
    SkillOpt-lite & 38.50 & 1.450 & 52.80 & 2.900 & 11.40 & 0.520 & 10.10 & 1.150 & 131.60 & 5.950 & 25.60 & 1.150 \\
    EvoSkill & 37.20 & 1.410 & 51.20 & 2.820 & 11.10 & 0.510 & 9.89 & 1.110 & 128.30 & 5.830 & 24.80 & 1.110 \\
    \rowcolor{oursgray}
    \textbf{CDSE} & 37.00 & 1.400 & 50.90 & 2.800 & 11.00 & 0.500 & 9.84 & 1.100 & 127.80 & 5.800 & 24.70 & 1.100 \\
    \bottomrule
  \end{tabularx}
  \caption{Input and output token consumption (millions of tokens) in a single evolution round with GLM-5.2 and Claude Code.}
  \label{tab:single-round-tokens}
\end{table*}

Table~\ref{tab:full-token-efficiency} extends the main-text input-token
efficiency comparison to all six benchmarks and includes output tokens.
Each value is the corresponding single-round token count divided by the
score gain over Base in the same round. Nonpositive gains are marked N/A.

\begin{table*}[!t]
  \centering
  \footnotesize
  \setlength{\tabcolsep}{3pt}
  \begin{tabularx}{\textwidth}{@{}l*{12}{>{\centering\arraybackslash}X}@{}}
    \toprule
    & \multicolumn{2}{c}{OfficeQA}
    & \multicolumn{2}{c}{\shortstack{Spreadsheet\\Bench2}}
    & \multicolumn{2}{c}{SealQA}
    & \multicolumn{2}{c}{LiveMathBench}
    & \multicolumn{2}{c}{SWE-Pro}
    & \multicolumn{2}{c}{\shortstack{Automation\\Bench}} \\
    \cmidrule(lr){2-3}\cmidrule(lr){4-5}\cmidrule(lr){6-7}
    \cmidrule(lr){8-9}\cmidrule(lr){10-11}\cmidrule(lr){12-13}
    Method & Input & Output & Input & Output & Input & Output
    & Input & Output & Input & Output & Input & Output \\
    \midrule
    Trace2Skill & 12.34 & 0.466 & 15.41 & 0.844 & 5.89 & 0.267 & N/A & N/A & 41.30 & 1.867 & 23.90 & 1.050 \\
    SkillOpt-lite & \textbf{6.64} & \textbf{0.250} & 29.33 & 1.611 & 3.17 & 0.144 & 11.22 & 1.278 & 59.82 & 2.705 & 6.56 & 0.295 \\
    EvoSkill & 10.94 & 0.415 & 42.67 & 2.350 & 1.61 & 0.074 & 6.59 & 0.740 & 75.47 & 3.429 & 5.06 & 0.227 \\
    \rowcolor{oursgray}
    \textbf{CDSE} & 7.87 & 0.298 & \textbf{7.60} & \textbf{0.418} & \textbf{1.55} & \textbf{0.070} & \textbf{3.08} & \textbf{0.344} & \textbf{15.04} & \textbf{0.682} & \textbf{2.19} & \textbf{0.097} \\
    \bottomrule
  \end{tabularx}
  \caption{Single-round input and output tokens per percentage point of
  score gain over Base on all six benchmarks (million tokens/pp,
  $\downarrow$), using GLM-5.2 with Claude Code. Bold marks column minima
  before rounding. N/A indicates no positive gain: Trace2Skill decreases
  the LiveMathBench score by 1.4 points.}
  \label{tab:full-token-efficiency}
\end{table*}

\section{Extended Component Analysis}
\label{app:component-analysis}

Table~\ref{tab:component-invocation-turns} reports the mean invocation
turns underlying Figure~\ref{fig:skill-specialization}. These values
describe when a component is invoked; components that are never invoked
have no corresponding mean turn.

\begin{table}[t]
  \centering
  \small
  \setlength{\tabcolsep}{3pt}
  \renewcommand{\arraystretch}{1.12}
  \begin{tabular}{lrrrr}
    \toprule
    Benchmark & Meta & Exp. & Tool & Script \\
    \midrule
    OfficeQA & 1.60 & 1.10 & 1.80 & 3.60 \\
    SpreadsheetBench2 & 1.40 & 2.70 & 4.90 & 12.30 \\
    SealQA & 1.00 & 1.13 & 2.24 & 2.67 \\
    LiveMathBench & 6.25 & 2.76 & -- & -- \\
    SWE-Pro & 1.30 & 1.70 & 3.80 & 10.20 \\
    AutomationBench & 1.10 & 1.10 & 1.10 & 1.80 \\
    \bottomrule
  \end{tabular}
  \caption{Mean first invocation turn of the four skill components in Figure~\ref{fig:skill-specialization}, evaluated with GLM-5.2 under Claude Code. A dash indicates no invocation.}
  \label{tab:component-invocation-turns}
\end{table}

The components exhibit distinct but coordinated behavioral roles across tasks. Their removal 
effects vary substantially by benchmark, indicating that different tasks rely on different 
forms of reusable knowledge. AutomationBench provides a clear example of coordination: 
removing Meta, Experience, Tool, or Script reduces performance by 5.6, 8.2, 6.3, and 5.0 points, 
respectively, while every component is invoked in at least 76.7\% of episodes. In contrast, 
other benchmarks show more selective dependence. For example, SealQA relies primarily on Meta 
and Experience, whereas SWE-Pro is substantially more sensitive to Meta and Tool. These 
heterogeneous contribution profiles support the functional differentiation of the skill state 
rather than treating all reusable knowledge as a homogeneous collection of guidance.

Invocation frequency does not directly reflect functional importance. On SpreadsheetBench2, 
Script is invoked in only 6.0\% of episodes, yet removing it causes the largest performance 
drop on the benchmark, at 8.0 points. This suggests that infrequently activated components 
can still be critical when triggered at the appropriate stage of execution. Invocation timing 
further reveals task-conditioned specialization. Meta and Experience are generally selected 
early, while Script first appears much later on SpreadsheetBench2 and SWE-Pro, at turns 12.3 
and 10.2 on average. By contrast, Tool and Script are invoked almost immediately on 
AutomationBench. Together, these patterns show that the agent does not follow a fixed 
component pipeline, but adaptively selects and coordinates differentiated knowledge according 
to task demands. This behavior is consistent with the cognitive-science motivation of unity 
and diversity: distinct functional components remain separable in their roles while cooperating 
within a unified capability state.

\section{Additional Evolution Results}
\label{app:evolution-results}

\begin{figure*}[!t]
  \centering
  \includegraphics[width=\textwidth]{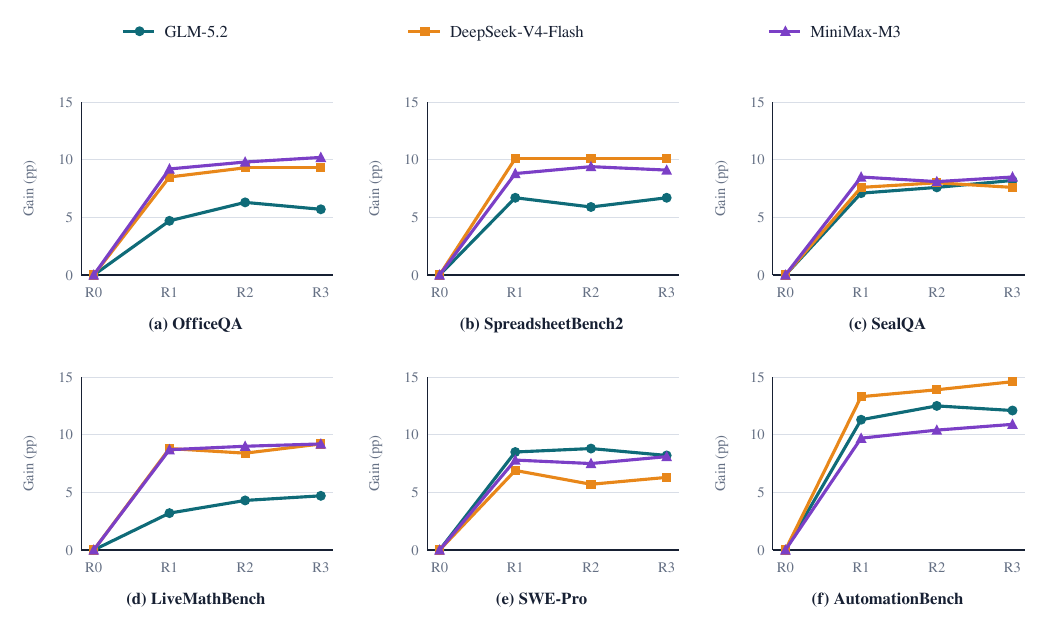}
  \caption{Evolution over three rounds on all six benchmarks with GLM-5.2,
  DeepSeek-V4-Flash, and MiniMax-M3 under Claude Code. Scores are shown as
  percentage-point gains over each model's no-skill baseline (R0). Each
  curve shows one reported score at each round; the same trajectory corpus
  is reused across rounds. The first round provides most of the gain, with
  smaller, benchmark-dependent changes thereafter.}
  \label{fig:evolution-three-models}
\end{figure*}

Figure~\ref{fig:evolution-rounds} in the main text reports three evolution
rounds on SpreadsheetBench2, LiveMathBench, and AutomationBench for GLM-5.2
and DeepSeek-V4-Flash under Claude Code, using the same trajectory corpus
across rounds. For GLM-5.2, R1 improves over the no-skill R0 baseline by
6.7, 3.2, and 11.3 percentage points on these benchmarks, respectively;
DeepSeek-V4-Flash gains 10.1, 8.8, and 13.3 points. At R3, the gains are
6.7, 4.7, and 12.1 points for GLM-5.2, and 10.1, 9.2, and 14.6 points for
DeepSeek-V4-Flash. Intermediate checkpoints show modest non-monotonic
fluctuations. Figure~\ref{fig:evolution-three-models}
compares all six benchmarks across GLM-5.2, DeepSeek-V4-Flash, and
MiniMax-M3 under Claude Code. Across these settings, R1 accounts for most
of the improvement, while later rounds yield smaller, task-dependent
changes. Tables~\ref{tab:evolution-glm}, \ref{tab:evolution-deepseek}, and
\ref{tab:evolution-minimax} report the corresponding R0--R3 scores for
each model.

Under this fixed-corpus setting, subsequent updates mainly produce small
revisions to the existing skill state. Sustained improvement may therefore
require additional execution experience that supplies new behavioral
evidence.

\begin{table}[H]
  \centering
  \small
  \setlength{\tabcolsep}{4pt}
  \renewcommand{\arraystretch}{1.12}
  \begin{tabular}{lrrrr}
    \toprule
    Benchmark & R0 & R1 & R2 & R3 \\
    \midrule
    OfficeQA & 71.5 & 76.2 & 77.8 & 77.2 \\
    SpreadsheetBench2 & 24.0 & 30.7 & 29.9 & 30.7 \\
    SealQA & 45.9 & 53.0 & 53.5 & 54.1 \\
    LiveMathBench & 83.0 & 86.2 & 87.3 & 87.7 \\
    SWE-Pro & 60.2 & 68.7 & 69.0 & 68.4 \\
    AutomationBench & 20.8 & 32.1 & 33.3 & 32.9 \\
    \bottomrule
  \end{tabular}
  \caption{Multi-round evolution scores (\%) for GLM-5.2 under Claude Code. R0 is the no-skill baseline.}
  \label{tab:evolution-glm}
\end{table}

\begin{table}[H]
  \centering
  \small
  \setlength{\tabcolsep}{4pt}
  \renewcommand{\arraystretch}{1.12}
  \begin{tabular}{lrrrr}
    \toprule
    Benchmark & R0 & R1 & R2 & R3 \\
    \midrule
    OfficeQA & 68.0 & 76.5 & 77.3 & 77.3 \\
    SpreadsheetBench2 & 16.8 & 26.9 & 26.9 & 26.9 \\
    SealQA & 51.8 & 59.4 & 59.8 & 59.4 \\
    LiveMathBench & 59.9 & 68.7 & 68.3 & 69.1 \\
    SWE-Pro & 57.8 & 64.7 & 63.5 & 64.1 \\
    AutomationBench & 14.3 & 27.6 & 28.2 & 28.9 \\
    \bottomrule
  \end{tabular}
  \caption{Multi-round evolution scores (\%) for DeepSeek-V4-Flash under Claude Code. R0 is the no-skill baseline.}
  \label{tab:evolution-deepseek}
\end{table}

\begin{table}[H]
  \centering
  \small
  \setlength{\tabcolsep}{4pt}
  \renewcommand{\arraystretch}{1.12}
  \begin{tabular}{lrrrr}
    \toprule
    Benchmark & R0 & R1 & R2 & R3 \\
    \midrule
    OfficeQA & 71.6 & 80.8 & 81.4 & 81.8 \\
    SpreadsheetBench2 & 25.3 & 34.1 & 34.7 & 34.4 \\
    SealQA & 49.4 & 57.9 & 57.5 & 57.9 \\
    LiveMathBench & 73.5 & 82.2 & 82.5 & 82.7 \\
    SWE-Pro & 57.2 & 65.0 & 64.7 & 65.3 \\
    AutomationBench & 9.1 & 18.8 & 19.5 & 20.0 \\
    \bottomrule
  \end{tabular}
  \caption{Multi-round evolution scores (\%) for MiniMax-M3 under Claude Code. R0 is the no-skill baseline.}
  \label{tab:evolution-minimax}
\end{table}

\section{Transfer Evaluation Details}
\label{app:transfer-details}

\paragraph{Cross-benchmark component transfer.}
We take an individual component evolved on SpreadsheetBench2 or
AutomationBench and make it available on a different target benchmark.
Table~\ref{tab:cross-benchmark-transfer} reports the score change from the
target Base on GLM-5.2. The parenthesized value is the fraction of target
episodes that explicitly load the transferred component. Under progressive
disclosure, component descriptions remain visible to the agent before
loading and may affect execution even when the full component is never
loaded. Invocation rate therefore measures explicit component loading,
rather than total exposure to the component.

The observed transfer differences are consistent with the functional
scopes of the components. Meta stores task-level control policies that can
remain useful across heterogeneous tasks. Experience stores
context-conditioned heuristics, while Tool and Script store
capability-specific instructions and executable procedures, respectively.
These forms of knowledge require progressively closer alignment between
source and target settings, consistent with their more selective or limited
cross-benchmark transfer.

\paragraph{Cross-model and cross-harness transfer.}
For cross-model transfer, we exchange complete skill states evolved by
GLM-5.2 and DeepSeek-V4-Flash while keeping Claude Code fixed. For
cross-harness transfer, we exchange states evolved under Claude Code and
Codex while keeping GLM-5.2 fixed. Each direction is evaluated on all six
benchmarks, yielding 12 receiving settings for each transfer experiment.
In Figure~\ref{fig:cross-model-harness-transfer}, Native denotes a state
evolved in the receiving configuration, whereas Transferred denotes one
evolved in the paired source configuration. Tables~\ref{tab:cross-model-scores}
and \ref{tab:cross-harness-scores} give the exact scores for these
settings, including the receiving Base and both skill states.

\begin{table}[H]
  \centering
  \footnotesize
  \setlength{\tabcolsep}{2pt}
  \renewcommand{\arraystretch}{1.12}
  \begin{tabular}{lrrrrrr}
    \toprule
    & \multicolumn{3}{c}{GLM-5.2} & \multicolumn{3}{c}{DeepSeek-V4-Flash} \\
    \cmidrule(lr){2-4}\cmidrule(lr){5-7}
    Benchmark & Base & N & T & Base & N & T \\
    \midrule
    OfficeQA & 71.5 & 77.2 & 72.5 & 68.0 & 77.3 & 72.5 \\
    SpreadsheetBench2 & 24.0 & 30.7 & 30.0 & 16.8 & 26.9 & 20.7 \\
    SealQA & 45.9 & 54.1 & 49.4 & 51.8 & 59.4 & 48.5 \\
    LiveMathBench & 83.0 & 87.7 & 85.8 & 59.9 & 69.1 & 71.6 \\
    SWE-Pro & 60.2 & 68.4 & 63.5 & 57.8 & 64.1 & 60.4 \\
    AutomationBench & 20.8 & 32.9 & 29.2 & 14.3 & 28.9 & 20.1 \\
    \bottomrule
  \end{tabular}
  \caption{Cross-model transfer scores (\%) for Figure~\ref{fig:cross-model-harness-transfer} (Claude Code fixed). Column groups identify the receiving configuration. N: Native; T: Transferred from the other configuration. Gains are relative to Base.}
  \label{tab:cross-model-scores}
\end{table}

\begin{table}[H]
  \centering
  \footnotesize
  \setlength{\tabcolsep}{2pt}
  \renewcommand{\arraystretch}{1.12}
  \begin{tabular}{lrrrrrr}
    \toprule
    & \multicolumn{3}{c}{Claude Code} & \multicolumn{3}{c}{Codex} \\
    \cmidrule(lr){2-4}\cmidrule(lr){5-7}
    Benchmark & Base & N & T & Base & N & T \\
    \midrule
    OfficeQA & 71.5 & 77.2 & 75.3 & 68.8 & 74.9 & 74.3 \\
    SpreadsheetBench2 & 24.0 & 30.7 & 28.6 & 22.7 & 31.3 & 26.1 \\
    SealQA & 45.9 & 54.1 & 47.5 & 22.6 & 27.6 & 25.5 \\
    LiveMathBench & 83.0 & 87.7 & 84.2 & 79.0 & 83.5 & 80.2 \\
    SWE-Pro & 60.2 & 68.4 & 64.2 & 62.3 & 68.9 & 65.6 \\
    AutomationBench & 20.8 & 32.9 & 25.1 & 42.2 & 47.1 & 46.0 \\
    \bottomrule
  \end{tabular}
  \caption{Cross-harness transfer scores (\%) for Figure~\ref{fig:cross-model-harness-transfer} (GLM-5.2 fixed). Column groups identify the receiving configuration. N: Native; T: Transferred from the other configuration. Gains are relative to Base.}
  \label{tab:cross-harness-scores}
\end{table}

Native evolution is stronger on average in both experiments. The positive
transfer gains indicate that substantial learned behavioral knowledge
remains useful after a change in model or execution harness, while the
advantage of native states indicates additional benefit from adapting to
the receiving configuration.

\section{Mechanism Ablation Details}
\label{app:mechanism-ablations}

We evaluate four mechanism ablations on SpreadsheetBench2, SWE-Pro,
and AutomationBench using GLM-5.2 under Claude Code. Unlike the
component-removal analysis in Figure~\ref{fig:skill-specialization},
these ablations modify the organization, maintenance, coordination,
or invocation of the skill state.

\paragraph{Without cognitive differentiation.}
The four components are merged into a unified skill maintained by
a generic update mechanism, using the same semantic gradients as
full CDSE.

\paragraph{Without specialized updates.}
The four-component structure and adaptive invocation are preserved,
but component-specific maintenance operators are replaced by a
shared update strategy.

\paragraph{Without cross-component coordination.}
Component-specific updates and validation are retained, while
cross-component conflict resolution, deduplication, and ownership
reconciliation are disabled.

\paragraph{Without the Adaptive Invocation Interface.}
This variant removes the content-aware invocation interface that describes
each component's scope, when it is useful, and its expected benefit to guide
component selection. Component-specific maintenance and cross-component
coordination are retained. The ablation therefore evaluates the contribution
of this interface to component selection during execution.
\section{Functional Composition of the Top-500 Skills}
\label{app:skill-survey}

\subsection{Survey Procedure and Annotation Labels}

We analyze the top-500 skill packages from the all-time ranking on
\url{https://skills.sh}. Human experts first examine the top-30 packages to
identify recurring functional content and establish 15 annotation labels:
14 labels grouped under Meta, Experience, Tool, and Script, and an auxiliary
label, O1 (Other / Non-behavioral / Unclear). We use GPT 6-Astra to apply
these expert-derived labels to the top-500 packages. Labels are non-exclusive, allowing a
package to contain multiple labels within and across components.

Table~\ref{tab:survey-labels} lists the labels and their frequencies. Meta
captures global task-level control, while Experience captures contextual
decisions and recovery; M3 and E4 distinguish global evidence control from
local interpretation or verification. Tool covers capability-specific
operational knowledge, and Script covers executable procedures (S1) and
their execution or validation contracts (S2). Script-related content may
therefore include contracts without executable code.

A package is counted once for a component if it contains any of that
component's labels. All percentages use the 500 packages as the denominator.
O1 is an auxiliary label and is excluded from the four-component count.

\paragraph{Validation against expert annotations.}
Human experts annotate 100 skill packages randomly sampled from the
500-package collection. Using these annotations as reference, we evaluate
GPT 6-Astra's labels separately for Meta, Experience, Tool, and Script.
Table~\ref{tab:annotation-validation} reports precision, recall, and
$F_1=2PR/(P+R)$. The macro average weights the four components equally.

\begin{table}[H]
  \centering
  \small
  \setlength{\tabcolsep}{5pt}
  \begin{tabularx}{\columnwidth}{@{}l*{3}{>{\centering\arraybackslash}X}@{}}
    \toprule
    Component & Precision & Recall & F1 \\
    \midrule
    Meta & 0.860 & 0.840 & 0.850 \\
    Experience & 0.880 & 0.840 & 0.860 \\
    Tool & 0.900 & 0.860 & 0.880 \\
    Script & 0.920 & 0.900 & 0.910 \\
    \midrule
    Macro average & 0.890 & 0.860 & 0.875 \\
    \bottomrule
  \end{tabularx}
  \caption{LLM annotation quality against expert labels on a random
  100-package subset. F1 is computed from the reported precision and recall.}
  \label{tab:annotation-validation}
\end{table}

\begin{table*}[!t]
\centering
\small
\begin{tabular}{lllrr}
\toprule
Component & ID & Label & Count & Percentage \\
\midrule
Meta & M1 & Task Planning \& Decomposition & 356 & 71.2\% \\
Meta & M2 & Progress \& Resource Control & 267 & 53.4\% \\
Meta & M3 & Global Verification \& Evidence Control & 214 & 42.8\% \\
Meta & M4 & Stopping \& Finalization & 345 & 69.0\% \\
Experience & E1 & Context-Conditioned Heuristic & 486 & 97.2\% \\
Experience & E2 & Failure Pattern \& Anti-Pattern & 421 & 84.2\% \\
Experience & E3 & Recovery \& Correction Strategy & 268 & 53.6\% \\
Experience & E4 & Local Interpretation \& Verification & 435 & 87.0\% \\
Tool & T1 & Tool Selection \& Applicability & 389 & 77.8\% \\
Tool & T2 & Tool Operation \& Sequencing & 377 & 75.4\% \\
Tool & T3 & Tool Output \& Limitation Handling & 286 & 57.2\% \\
Tool & T4 & Tool Failure \& Environment Handling & 257 & 51.4\% \\
Script & S1 & Reusable Executable Procedure & 98 & 19.6\% \\
Script & S2 & Execution Contract \& Validation & 68 & 13.6\% \\
Other & O1 & Other / Non-behavioral / Unclear & 19 & 3.8\% \\
\bottomrule
\end{tabular}
\caption{Fine-grained label prevalence among 500 skills. Labels are
non-exclusive; each count is the number of skills containing that label.}
\label{tab:survey-labels}
\end{table*}

\subsection{Survey Results}

Table~\ref{tab:survey-components} reports component prevalence. Experience
appears in 493 packages (98.6\%), Meta in 443 (88.6\%), Tool in 398 (79.6\%),
and Script-related content in 119 (23.8\%).

\begin{table}[!htbp]
\centering
\begin{tabular}{lrr}
\toprule
Component & Count / 500 & Percentage \\
\midrule
Meta & 443 & 88.6\% \\
Experience & 493 & 98.6\% \\
Tool & 398 & 79.6\% \\
Script & 119 & 23.8\% \\
\bottomrule
\end{tabular}
\caption{Component prevalence. A skill is counted once for a component if it
contains any of that component's labels.}
\label{tab:survey-components}
\end{table}

Table~\ref{tab:survey-counts} shows that 484 packages (96.8\%) contain at
least two components, including 380 (76.0\%) with at least three and 89
(17.8\%) with all four. Only 16 packages (3.2\%) contain a single component.
This prevalence of multiple functional components motivates CDSE's
functionally differentiated skill organization.

\begin{table}[!htbp]
\centering
\begin{tabular}{lrr}
\toprule
Components & Skills & Percentage \\
\midrule
0 & 0 & 0.0\% \\
1 & 16 & 3.2\% \\
2 & 104 & 20.8\% \\
3 & 291 & 58.2\% \\
4 & 89 & 17.8\% \\
\midrule
Total & 500 & 100.0\% \\
\bottomrule
\end{tabular}
\caption{Distribution of the number of functional components per skill.
Unlike component marginals, these categories are mutually exclusive.}
\label{tab:survey-counts}
\end{table}

These descriptive results characterize the selected highly ranked packages.

\section{Implementation Details of CDSE}
\label{app:method-details}

This section specifies the implementation contracts underlying the three
stages in Section~\ref{sec:method}. We use the same component state and
trajectory notation as the main text. The following details expand the
filtering, evidence selection, and validation steps summarized there.

\subsection{Trajectory Filtering and Finding Extraction}
\label{detail:sec:finding-induction}

\paragraph{Preserving evidence while filtering logs.}
The input to discovery is the evaluated corpus $\mathcal{D}^{(t)}$.
A deterministic filter produces $\bar{\tau}_i=P(\tau_i)$ by removing runtime-only
heartbeats, retry notifications, lifecycle status, duplicated terminal metrics,
and redundant metadata and wrappers. It retains ordered reasoning, tool calls
and observations, permission failures, final responses, and termination state,
while retaining the raw trace separately. Analysis operates on filtered
execution evidence rather than a model-generated summary.

\paragraph{Outcome-conditioned behavioral diagnosis.}
The analyzer uses $(\bar{\tau}_i,y_i)$ to identify a reusable direction of
behavioral change, or \emph{semantic gradient}. Natural-language update
directions follow the semantic and textual gradient perspective
\citep{shinn2023reflexion,pryzant2023automatic}.
Here, outcome-contrastive discovery applies different diagnostic objectives to
successes and failures; it requires neither matched trajectory pairs nor a
numerical gradient. For a success, the analyzer reconstructs a minimal
trace-supported account of how the result was reached and rates its integrity
as valid, fragile, invalid, or unknown. It extracts behavior worth preserving
and safeguards that would strengthen a fragile success. For a failure, it
locates the earliest clearly evidenced decisive error, missing transition or
recovery, or external obstacle in the inspected trace, together with the
smallest supported recovery. Answer mismatch, timeout, and parse failure are
symptoms whose causes must be supported by the trace. A successful outcome
alone does not establish a sound procedure, and unresolved causes remain
unknown.

\paragraph{Functional abstraction into semantic gradients.}
Let $z_i=s(y_i)\in\{+,-\}$ map the task-specific outcome to its
success/failure label.
The two diagnostic paths feed the same functional abstraction step:
\begin{equation}
  \mathcal{F}_i=A_{z_i}(P(\tau_i),y_i),\qquad
  f_{ij}=(g_{ij},e_{ij}),
  \label{detail:eq:gradient-discovery}
\end{equation}
where $g_{ij}$ expresses a reusable behavioral update and $e_{ij}$ identifies
its supporting trace lines. $A_+$ seeks behavior to preserve or strengthen;
$A_-$ seeks an evidenced failure and supported recovery. Each analysis emits a
core finding, up to two reusable lessons, and at most one script suggestion
for a repeated deterministic operation:
\begin{equation}
  \mathcal{F}_i=\{f_{ij}\}_{j=1}^{m_i},\qquad
  1\leq m_i\leq 4,
  \quad
  \mathcal{F}=\bigcup_{i=1}^{N_t}\mathcal{F}_i.
  \label{detail:eq:finding-set}
\end{equation}
Each candidate carries exact trace-line provenance. The analyzer separates
observed evidence from reusable generalization; it does not reconstruct
unavailable ground-truth answers. The output $\mathcal{F}$ contains candidate semantic gradients; the next
stage determines which are worth retaining and what function each should serve.

\subsection{Retention and Attribution Rules}
\label{detail:sec:gradient-consolidation}

Consolidation maps the semantic gradients $\mathcal{F}$ to component-specific
update evidence $\{\mathcal{F}_k\}_{k\in\mathcal{K}}$. The distinction between
knowledge structures and control processes motivates separating a finding's
content from its execution role \citep{atkinson1968human}. Evidence for
dissociable forms of knowing further motivates specialized maintenance
\citep{cohen1980preserved}. CDSE instantiates these principles through a
retention gate followed by functional role attribution.

\paragraph{Retention gate.}
The router first tests whether a candidate is supported, actionable, reusable,
and likely to change execution. It drops generic advice, unsupported diagnoses,
sample-specific answers, and redundant core analyses already covered by the
extracted lessons. For failures, dropping a redundant core requires the lessons
to preserve both the diagnosis and the recovery. Writing the gate as $r(f)\in\{0,1\}$, its output is
$\mathcal{F}^{\mathrm{keep}}=\{f\in\mathcal{F}:r(f)=1\}$.
Only retained gradients proceed to attribution.

\paragraph{Functional role attribution.}
For each retained gradient, the router identifies a primary component and
at most one secondary component. The set $\rho(f)$ records membership in
these targets, with primary status stored separately:
\begin{equation}
  \begin{aligned}
  \rho(f)&\subseteq\mathcal{K},\quad 1\leq|\rho(f)|\leq2,\\
  \mathcal{F}_k&=\{f\in\mathcal{F}^{\mathrm{keep}}:k\in\rho(f)\}.
  \end{aligned}
  \label{detail:eq:routing}
\end{equation}
A secondary component is included only when the same evidence has substantial
value under another functional responsibility. Meta
receives task-wide control changes; Experience receives heuristics activated
by identifiable local conditions. Tool receives operational guidance whose
validity depends on a named tool or capability; Script receives proposals for
deterministic, testable procedures or their execution contracts. For example, checking units before comparing
values is contextual experience, whereas a restriction specific to a spreadsheet
library belongs to Tool. A repeatable transformation with explicit inputs and
tests can instead become a Script proposal. The router neither rewrites the
findings nor generates skill content, preserving the distinction between
evidence, ownership, and the subsequent update.

\subsection{Component Maintenance and Validation}
\label{detail:sec:typed-evolution}

Once ownership is established, each component evolves through its own operator:
\begin{equation}
  \widehat{\mathcal{B}}^{(t+1)}_k =
  U_k\!\left(\mathcal{B}^{(t)}_k,\mathcal{F}_k\right),
  \qquad k\in\mathcal{K}.
  \label{detail:eq:type-update}
\end{equation}
Each $U_k$ produces a component-level candidate. The full operator
$\mathcal{U}$ subsequently coordinates and validates these candidates before
returning the next skill state. Specialized maintenance followed by integration reflects the cognitive
principle that differentiated functions can remain coordinated
\citep{miyake2000unity}.

\paragraph{Meta and Experience: consolidating behavioral rules.}
Both components store ordered atomic rules with stable identifiers. A finding
aligns with at most two existing rules as reinforcing, extending, or correcting
the same knowledge unit; it is split only into independently actionable parts.
Evidence for one rule is aggregated into one revision, while unaffected rules
are preserved. Unaligned findings with compatible triggers, actions, scope, and
constraints induce new rules, with proposals reconciled across batches. When
capacity is exceeded, the updater shortens rules, merges genuine redundancy,
and finally removes low-value rules. Meta applies this process to task-level
control, whereas Experience requires a local activation condition. Thus, shared
rule machinery maintains distinct behavioral responsibilities.

\paragraph{Tool: maintaining capability-grounded knowledge.}
Tool organizes operational rules within stable capability categories. Findings
first align with an existing tool or category; unmatched category proposals are
reconciled across batches. Rule alignment, revision, and induction then occur
within each category. Local capacity control prevents frequently encountered
tools from displacing knowledge about other capabilities.

\paragraph{Script: turning suitable feedback into verified procedures.}
Script findings first align with existing executable components. Residual
proposals pass a worthiness gate requiring repeatable deterministic behavior,
explicit I/O and error contracts, available inputs, feasible synthetic tests,
and material savings over one-off tool use. Compatible proposals are grouped;
others are deferred. Each work item selects a documentation update, local
modification, replacement, creation, or deferral. Implementations stay within
allocated paths, declare dependencies, and pass self-tests. An independent
verifier then checks the assembled Script component in a clean environment using component
and full-suite tests. One bounded repair is allowed; unverified code is withheld.
This branch gives procedural changes an executable acceptance contract.

\paragraph{Cross-component coordination.}
\label{detail:sec:coordination}
Component updates produce a candidate skill state
$\widehat{\mathcal{B}}^{(t+1)}$. A coordinator $C$ reads all four components and
makes minimal edits to resolve concrete conflicts, duplication, ownership
ambiguity, and documentation mismatches. Script artifacts remain authoritative.
A complete Meta, Experience, or Tool duplicate can be removed when its meaning
survives in the appropriate component; partial overlap is clarified. The
coordinator cannot introduce, renumber, or reorder rules, change Tool category
membership, or alter Script entries or executables. A validator compares the
candidate with its input snapshot and confines changes to the four component
specifications, preserving the previously verified programs.

\paragraph{Adaptive invocation and publication.}
A description rewriter $D$ derives each component's scope, timing, and expected
benefit from the coordinated content. Only the descriptions visible before
component loading may change; a validator requires well-formed, distinct
descriptions and byte-identical remaining content and artifacts. Successful
coordination and description validation produce
$\mathcal{B}^{(t+1)}=D(C(\widehat{\mathcal{B}}^{(t+1)}))$;
$\mathcal{U}$ is a partial operator, undefined when either stage fails
validation, in which case no updated state is published. These checks enforce edit
contracts, while task performance is assessed separately in our experiments.
For execution history $h_\ell$ at step $\ell$, write the agent's invocation
choice as
\begin{equation}
  I_\ell=\operatorname{Select}_{\pi_\theta}
  \!\left(h_\ell,\{d_k^{(t+1)}\}_{k\in\mathcal{K}}\right)
  \subseteq\mathcal{K},
  \label{detail:eq:adaptive-invocation}
\end{equation}
where $d_k^{(t+1)}$ is the revised description of component $k$ and $I_\ell$
contains the components selected for loading at that step. Selection uses the agent's ordinary invocation mechanism. As the task
progresses, different components become relevant; their use changes agent
behavior and supplies execution evidence for subsequent skill evolution.

\section{Experimental Setup}
\label{app:experimental-details}

\subsection{Benchmarks and Data Splits}
\label{app:benchmark-details}

Table~\ref{tab:benchmark-selections} lists the task families and dataset
selections used in our evaluation. We split each resulting benchmark
collection into training and test sets at a 2:7 ratio. The training split
supplies the trajectories used for skill evolution, and all reported results
are computed on the test split.

\begin{table*}[!t]
  \centering
  \small
  \setlength{\tabcolsep}{5pt}
  \begin{tabularx}{\textwidth}{@{}l>{\raggedright\arraybackslash}p{4.0cm}>{\raggedright\arraybackslash}X@{}}
    \toprule
    Benchmark & Task family & Dataset selection \\
    \midrule
    OfficeQA & Document-grounded question answering & Easy and Hard subsets \\
    SpreadsheetBench2 & Spreadsheet manipulation & \texttt{Financial\_Model} and \texttt{Template} subsets \\
    SealQA & Long-context evidence synthesis & Complete dataset \\
    LiveMathBench & Mathematical reasoning & \texttt{202605} and \texttt{202606} subsets \\
    SWE-Pro & Repository-level software engineering & Three Python subsets: \texttt{pandas}, \texttt{scikit-learn}, and \texttt{xarray} \\
    AutomationBench & Tool-based business workflows & All instances from the Sales and Marketing domains \\
    \bottomrule
  \end{tabularx}
  \caption{Task families and dataset selections. Each selected collection is split into training and test sets at a 2:7 ratio.}
  \label{tab:benchmark-selections}
\end{table*}

\subsection{Common Evaluation Protocol}
\label{app:evaluation-protocol}

We test GLM-5.2, DeepSeek-V4-Flash, and MiniMax-M3 under both Claude Code
and Codex. Within each model--harness block, all methods receive access to
the same underlying training split, task instances, tool access, and context
limit. Each question is subject to benchmark-specific maximum runtime and
turn limits. If execution reaches either limit without completing the task,
the outcome is counted as a failure. Base is the same agent without an
evolved skill state. CDSE starts from the same skill initialization and updates one unified
state whose four functional components are selectively loaded at inference
time. Table~\ref{tab:main-results} reports the state after three evolution rounds
for each skill-evolution method.

We apply the benchmark-specific scoring procedures described below. Reported
differences are
percentage-point changes within a benchmark. For aggregate statements, we
macro-average these changes over the relevant model--harness--benchmark
cells; we do not average the underlying heterogeneous task metrics.

\subsection{Scoring Rules and Failure Handling}
\label{app:benchmark-scoring}

Evaluation depends on the task output: OfficeQA and SealQA score final
answers, SpreadsheetBench2 scores generated workbooks, LiveMathBench scores
selected answer labels, SWE-Pro scores executable test outcomes, and
AutomationBench scores the final environment state. The following rules
specify the reported success rates and distinguish them from auxiliary
partial-credit scores.

\paragraph{OfficeQA.}
Predicted and reference answers undergo numeric extraction and normalization,
including magnitude units, signs, and thousands separators. Numeric answers
are compared at relative-error tolerances
$\mathcal{T}=\{0,0.01,0.025,0.05,0.1\}$, yielding
\[
 s = \frac{\sum_{t\in\mathcal{T}}w_t\,\mathbf{1}[\text{match at }t]}
               {\sum_{t\in\mathcal{T}}w_t},
 \qquad w_t=\frac{1}{1+20t}.
\]
A match also requires the reference's nonnumeric text to occur in the
prediction. For a reference containing multiple numbers, every number must
match. Integers from 1900 to 2100 are treated as year candidates and excluded
from numeric matching unless the reference itself is a year. Purely textual
answers use normalized substring or exact matching. Empty answers receive
zero and count as failures in the reported success rate. Strict success
requires $s\geq0.8$; the continuous mean score is an auxiliary diagnostic. In the Codex evaluation, timeouts and crashes may be
retried at most twice. In Claude Code, reaching the 60-turn limit is counted
as a failure without retry.

\paragraph{SpreadsheetBench2.}
Only the hidden answer regions of the workbook are scored. Comparing the
input workbook with the reference partitions these regions into unchanged
(regression) cells and cells requiring modification. The output is evaluated
separately on both groups. A task succeeds only when both accuracies are
1.0, after regression accuracy of at least 0.998 is rounded to 1.0; no
partial credit contributes to the reported task success rate. Numeric cells
allow 1\% relative error, with absolute tolerance 0.01 when either value is
zero. Numeric strings are parsed and rounded to two decimal places, and
dates are converted to Excel serial values. Formula comparisons ignore case
and absolute-reference markers. Array formulas use their formula text;
recognized financial placeholders are normalized, as are blank cells and
zero where specified by the evaluator. Excel error values trigger a fallback
to formula-text comparison. Missing outputs fail the task; missing sheets
or corrupted files invalidate the affected scoring regions.

\paragraph{SealQA.}
A shared offline judge labels each answer Correct, Incorrect, or Not
Attempted; only Correct receives one point. The answer must contain the
reference's essential information without contradictory claims or internal
inconsistency. Differences in wording, punctuation, capitalization, and
contextually redundant information are ignored. Uncertain phrasing is
allowed when the required information is complete and consistent; an
incomplete or unattempted answer receives zero. All evaluated models,
harnesses, and methods use the same few-shot judging template and the
\texttt{deepseek-v4-flash-huawei} judge with temperature 0 and a 2,048-token
output limit. Judging is performed after trajectories are saved.

\paragraph{LiveMathBench.}
Our multiple-choice evaluation uses a deterministic scorer. It extracts the
last \texttt{ANSWER:} declaration containing a letter from A to E and reverses
the deterministic option shuffle before comparing with the reference label.
Only the correct option receives one point. The designated weaker-true
option, other incorrect options, missing answer declarations, exhausted turn
budgets, and unresolved execution errors all receive zero. No language-model
judge is used.

\paragraph{SWE-Pro.}
An instance is resolved only if all tests in both the FAIL\_TO\_PASS and
PASS\_TO\_PASS sets pass after the agent's changes. This binary criterion
requires the intended fix while preserving previously passing behavior;
partial test coverage does not count as task success.

\paragraph{AutomationBench.}
Rubric assertions evaluate the final world state rather than the agent's
answer text. The reported metric is binary task completion: every scored
assertion must pass. The fraction of passing assertions provides auxiliary
partial credit. Assertions already satisfied initially and still satisfied
at the end are excluded to avoid rewarding inaction; initially satisfied
assertions that the agent breaks remain failures. Tasks whose correct
behavior is inaction can explicitly retain their negative assertions.
Negative assertions check that prohibited actions do not occur, and checks
marked as unscored or explicitly excluded do not contribute to the metric.

\subsection{Claude Code Invocation Parameters}
\label{app:claude-code-parameters}

The following launch templates specify the benchmark-dependent Claude Code
configuration used in our evaluation. Angle-bracketed entries denote the
question or task prompt, evaluated model, corpus path, prompt file, or MCP
configuration supplied for each run. The templates preserve the tool and
permission flags used in the experiments. The answer schema is shown in
full where specified; the source configuration does not provide the literal
benchmark prompts, Friday-search server JSON, or arm-specific tool lists.

\paragraph{OfficeQA}
The question is passed through \texttt{-p}, and \texttt{--add-dir} exposes the document corpus. The analyst system prompt is appended to the agent context. The enabled tools are Read, Glob, Grep, and Bash, with a 60-turn limit. The command explicitly selects GLM-5.2 and records safe mode with the \texttt{acceptEdits} permission mode. Verbose streaming JSON and the supplied schema require string-valued \texttt{final\_answer} and \texttt{reasoning} fields.

\begin{Verbatim}[fontsize=\scriptsize,breaklines=true,breakanywhere=true]
claude \
  -p "<question>" \
  --model glm-5.2 \
  --safe-mode \
  --output-format stream-json --verbose \
  --json-schema '{"type":"object","properties":{"final_answer":{"type":"string"},"reasoning":{"type":"string"}},"required":["final_answer","reasoning"]}' \
  --allowed-tools Read,Glob,Grep,Bash \
  --add-dir <document corpus> \
  --append-system-prompt "<analyst system prompt>" \
  --permission-mode acceptEdits \
  --max-turns 60
\end{Verbatim}

\paragraph{SpreadsheetBench2}
The task prompt is passed through \texttt{-p}, and the system prompt is loaded from a file. The tool set comprises Bash, Read, Write, Edit, Glob, Grep, and Skill; Agent, WebFetch, and WebSearch are explicitly disabled. Evaluation uses a 60-turn limit, \texttt{bypassPermissions}, and verbose JSON output. This launch template does not include an explicit \texttt{--model} argument.

\begin{Verbatim}[fontsize=\scriptsize,breaklines=true,breakanywhere=true]
claude \
  --max-turns 60 \
  --permission-mode bypassPermissions \
  --tools "Bash Read Write Edit Glob Grep Skill" \
  --disallowed-tools "Agent WebFetch WebSearch" \
  --append-system-prompt-file <system_prompt.md> \
  --output-format json --verbose \
  -p "<task prompt>"
\end{Verbatim}

\paragraph{SealQA}
The question and evaluated model are supplied explicitly. Search is provided by the Friday-search MCP server configuration; \texttt{--strict-mcp-config} restricts MCP loading to that configuration. The allowed-tool list is supplied for the corresponding experimental arm, while WebSearch and WebFetch are disabled. The system prompt may include the indicated nudge. The run allows 80 turns and uses \texttt{bypassPermissions}, verbose streaming JSON, and the same two-field answer schema as OfficeQA.

\begin{Verbatim}[fontsize=\scriptsize,breaklines=true,breakanywhere=true]
claude \
  --model "<model>" \
  --verbose \
  -p "<question>" \
  --output-format stream-json \
  --json-schema '{"type":"object","properties":{"final_answer":{"type":"string"},"reasoning":{"type":"string"}},"required":["final_answer","reasoning"]}' \
  --strict-mcp-config \
  --mcp-config '<friday-search server JSON>' \
  --permission-mode bypassPermissions \
  --allowed-tools "<arm-specific list>" \
  --disallowed-tools "WebSearch WebFetch" \
  --append-system-prompt "<SYSTEM_PROMPT [+ nudge]>" \
  --max-turns 80
\end{Verbatim}

\paragraph{LiveMathBench}
The question prompt, evaluated model, and system prompt are passed directly. Bash, Write, and Edit are enabled, and the turn limit is 15. The recorded command includes \texttt{acceptEdits}, \texttt{--dangerously-skip-permissions}, and \texttt{--safe-mode} together, with verbose streaming JSON output. These flags are reproduced as configured in the experiment.

\begin{Verbatim}[fontsize=\scriptsize,breaklines=true,breakanywhere=true]
claude -p <question_prompt> \
  --output-format stream-json --verbose \
  --model <model> \
  --append-system-prompt <SYSTEM_PROMPT> \
  --allowed-tools "Bash Write Edit" \
  --permission-mode acceptEdits \
  --dangerously-skip-permissions \
  --safe-mode \
  --max-turns 15
\end{Verbatim}

\paragraph{SWE-Pro}
The repository-level task prompt is supplied through \texttt{-p}. Bash, Read, Grep, Glob, Edit, and Write support repository inspection and modification. The command uses safe mode, \texttt{bypassPermissions}, verbose streaming JSON, and a 250-turn limit. As in SpreadsheetBench2, the supplied launch template omits an explicit \texttt{--model} argument.

\begin{Verbatim}[fontsize=\scriptsize,breaklines=true,breakanywhere=true]
claude -p "<task prompt>" \
  --safe-mode \
  --permission-mode bypassPermissions \
  --output-format stream-json \
  --verbose \
  --max-turns 250 \
  --allowed-tools "Bash Read Grep Glob Edit Write"
\end{Verbatim}

\paragraph{AutomationBench}
The user prompt is supplied through \texttt{--print}, with an explicit evaluated model. A per-task MCP configuration specifies the task tools, and strict MCP loading is enabled. The system prompt is read from a file. The built-in tool set and allowed-tool list both specify Skill and Bash. Evaluation uses a 60-turn limit, \texttt{bypassPermissions}, and verbose JSON output.

\begin{Verbatim}[fontsize=\scriptsize,breaklines=true,breakanywhere=true]
claude --print <user_prompt> \
  --model <model> \
  --mcp-config <per-task.mcp.json> --strict-mcp-config \
  --append-system-prompt-file <system_prompt.txt> \
  --output-format json --verbose \
  --max-turns 60 \
  --permission-mode bypassPermissions \
  --tools Skill,Bash --allowed-tools Skill,Bash
\end{Verbatim}

\subsection{Codex Invocation Parameters}
\label{app:codex-parameters}

The following launch templates list the recorded Codex configuration flags
for the same six benchmarks. Angle-bracketed entries denote the evaluated
model, system prompt, gateway URL, or per-task paths. All templates use
\texttt{codex exec}, skip the Git repository check, request JSON output, and
set \texttt{model\_context\_window=1000000}. We reproduce the supplied
flags without inferring settings omitted from the templates.

\paragraph{OfficeQA}
The command selects GLM-5.2, uses the \texttt{workspace-write} sandbox,
disables web search, and supplies an OpenAI-compatible gateway URL.
The system prompt is supplied through \texttt{developer\_instructions}.

\begin{Verbatim}[fontsize=\scriptsize,breaklines=true,breakanywhere=true]
codex exec --skip-git-repo-check --json \
  -c model="glm-5.2" \
  -s workspace-write \
  -c model_context_window=1000000 \
  -c tools.web_search=false \
  -c model_providers.codex.base_url="<OpenAI-compatible gateway URL>" \
  -c developer_instructions="<system prompt>"
\end{Verbatim}

\paragraph{SpreadsheetBench2}
The evaluated model is supplied explicitly. The command uses
\texttt{workspace-write}, disables web search, and loads
\texttt{developer\_instructions} from \texttt{system\_prompt.md}.

\begin{Verbatim}[fontsize=\scriptsize,breaklines=true,breakanywhere=true]
codex exec \
  --skip-git-repo-check \
  --json \
  -s workspace-write \
  -c model="<model>" \
  -c model_context_window=1000000 \
  -c tools.web_search=false \
  -c developer_instructions="$(cat system_prompt.md)"
\end{Verbatim}

\paragraph{SealQA}
The evaluated model and gateway URL are supplied explicitly. Unlike the
OfficeQA and SpreadsheetBench2 templates, this command enables web search.
It uses \texttt{workspace-write} and passes the system prompt through
\texttt{developer\_instructions}.

\begin{Verbatim}[fontsize=\scriptsize,breaklines=true,breakanywhere=true]
codex exec \
  --skip-git-repo-check \
  --json \
  -s workspace-write \
  -c model="<model>" \
  -c model_providers.codex.base_url="<gateway URL>" \
  -c model_context_window=1000000 \
  -c tools.web_search=true \
  -c developer_instructions="<SYSTEM_PROMPT>"
\end{Verbatim}

\paragraph{LiveMathBench}
The evaluated model is selected through \texttt{-m}. The command uses
\texttt{workspace-write}, disables web search, and supplies the system
prompt through \texttt{developer\_instructions}.

\begin{Verbatim}[fontsize=\scriptsize,breaklines=true,breakanywhere=true]
codex exec --skip-git-repo-check --json \
  -s workspace-write \
  -m <MODEL> \
  -c model_context_window=1000000 \
  -c tools.web_search=false \
  -c developer_instructions="<SYSTEM_PROMPT>"
\end{Verbatim}

\paragraph{SWE-Pro}
The command supplies the evaluated model and system prompt and uses
\texttt{workspace-write}. The recorded template does not include an
explicit web-search setting.

\begin{Verbatim}[fontsize=\scriptsize,breaklines=true,breakanywhere=true]
codex exec --skip-git-repo-check --json \
  -s workspace-write \
  -c model="<model>" \
  -c model_context_window=1000000 \
  -c developer_instructions="<system prompt>"
\end{Verbatim}

\paragraph{AutomationBench}
This template selects GLM-5.2 and uses the \texttt{danger-full-access}
sandbox. The \texttt{ab} MCP server is launched with the specified Python
interpreter, server script, task file, and final-world-state output file.
The system prompt is supplied through \texttt{developer\_instructions}.

\begin{Verbatim}[fontsize=\scriptsize,breaklines=true,breakanywhere=true]
codex exec \
  --skip-git-repo-check \
  --json \
  -s danger-full-access \
  -m glm-5.2 \
  -c model_context_window=1000000 \
  -c 'mcp_servers.ab.command="<abs-path-to-venv-python>"' \
  -c 'mcp_servers.ab.args=["<abs>/ab_mcp_server.py","--task-file","<abs>/task.json","--output-file","<abs>/world_final.json"]' \
  -c developer_instructions='"<system prompt>"'
\end{Verbatim}

\subsection{Baseline Configurations}
\label{app:baseline-configurations}

We follow the parameter settings recommended by the original methods unless
specified otherwise. The configurations used in our comparison are as follows.

\paragraph{Trace2Skill.}
For Trace2Skill~\citep{ni2026trace2skill}, we combine the Error and Success analysis branches and merge all generated updates.

\paragraph{EvoSkill.}
For EvoSkill~\citep{alzubi2026evoskill}, we run 1.5 epochs with a frontier size
of 3 and construct updates only from erroneous examples. For benchmarks
without binary outcomes, we identify errors using the corresponding
task-score threshold.

\paragraph{SkillOpt-lite.}
For SkillOpt-lite~\citep{yang2026skillopt}, we run 10 optimization rounds with
a batch size of 10.

\section{System Prompt for Gradient Consolidation}
\label{app:consolidation-system-prompt}

This appendix reproduces the system prompt for
\emph{Cognition-Guided Gradient Consolidation}, the second stage of CDSE
(Section~\ref{sec:gradient-consolidation}). The remaining prompts and
implementation code are provided in the \emph{Software} supplementary material.

\begin{Verbatim}[fontsize=\footnotesize,breaklines=true,breakanywhere=true,breaksymbolleft={}]
You are the routing classifier in a four-skill evolution pipeline.

You receive exactly one completed trajectory-analysis JSON. The original trajectory, other analyses, existing skills, and external resources are unavailable and must not be reconstructed. Classify only the supplied findings; do not rewrite, merge, split, summarize, or create findings.

## Targets

- `meta`: task-level control knowledge, including priorities, evidence-access boundaries, preservation of user constraints, phase transitions, search or turn budgets, stopping and finalization, resolution of known contradictions, and prevention of answer leakage or unsupported answers.
- `experience`: context-dependent heuristics, anti-patterns, verification practices, interpretation risks, and recovery tied to a particular data shape, source situation, or calculation context.
- `tool`: actionable knowledge about using an existing named environment tool, including selection, sequencing, output interpretation, permissions, limitations, misuse, inefficient repetition, and recovery.
- `script`: a proposed reusable deterministic operation with defined inputs and outputs that can be implemented and automatically tested.

## Decision order

Evaluate each finding in this order:

1. Does it control whether the task may proceed, return to an earlier phase, stop, or finalize? Consider `meta`.
2. Does it give actionable guidance about an existing named environment tool? Consider `tool`.
3. Does it propose a new deterministic executable capability? Consider `script`.
4. Is its remaining value a local heuristic, anti-pattern, verification practice, or recovery? Consider `experience`.

This order guides evaluation but does not restrict target combinations.

## Boundaries

Route by reusable function, not by words that merely appear in the finding.

For `meta`, judge semantic scope rather than observed frequency. A task-level rule may be routed to `meta` from one trajectory; cross-trajectory frequency and final admission are evaluated later.

Do not use `experience` as the default for every useful lesson. Use it for knowledge that remains local after checking `meta`, `tool`, and `script`.

For `tool`, require an explicit named tool and actionable guidance about how it should or should not be used. Bash, Read, Write, Edit, Glob, Grep, and other named environment tools count. Merely using or mentioning a tool is insufficient. Documents, datasets, source tables, file formats, benchmarks, and domain-specific data sources are not tools.

Prefer `tool` when the central rule changes if the named tool is replaced; prefer `experience` when the rule remains fully tool-independent. A concrete named-tool sequence, usage boundary, output-handling rule, or recovery has meaningful `tool` value even when it also supports a broader verification or efficiency practice. Use `tool` for guidance about an existing capability and `script` for a new maintained executable capability. A shell snippet or generic calculation is not a tool finding merely because Bash can run it.

For `script`, require a deterministic and testable operation. Do not route semantic judgment, source-authority selection, business-definition choice, general planning, or “find the correct answer” capabilities to `script`.

## Keep or drop

Drop findings that are generic, unsupported, purely sample-specific, or unlikely to change future behavior. Apply the core-specific redundancy rules below.

A valid success may contribute positive behavior. A fragile success should contribute only supported risks or improvements. An invalid success may contribute an anti-pattern or recovery, but not a positive pattern. Drop unknown-success claims whose causal basis is unverified.

Drop a `success_analysis` core when its reusable behavior, risk, or improvement is already captured by `reusable_lessons`.

A `failure_analysis` core may remain useful because it combines cause and recovery. Keep it when its failure mechanism, task-control implication, recovery, or routing value adds material information not fully preserved by the lessons. Drop it as redundant only when both its reusable cause and recovery are fully covered; partial overlap is insufficient. The drop reason must state that both are covered. Do not keep sample-specific detail alone.

## Output behavior

A kept finding must have one or at most two ordered targets. The first is primary. Use two targets when a finding has both a concrete tool-specific rule and a meaningful tool-independent lesson; order them by primary value. Otherwise add a secondary target only when it has substantial independent value.

Do not generate skill content or reason about downstream wording. Keep each routing reason to one concise sentence. Return only JSON matching the supplied schema.
\end{Verbatim}

\end{document}